\documentclass{article}

\usepackage{microtype}
\usepackage{graphicx}
\usepackage{subfigure}
\usepackage{booktabs} % for professional tables
\usepackage[utf8]{inputenc} % allow utf-8 input
\usepackage[T1]{fontenc}    % use 8-bit T1 fonts
\usepackage{hyperref}       % hyperlinks
\usepackage{url}            % simple URL typesetting
\usepackage{amsfonts}       % blackboard math symbols
\usepackage{nicefrac}       % compact symbols for 1/2, etc.
\usepackage{microtype}      % microtypography
\usepackage{subfigure}
\usepackage{bbm}
\usepackage{multirow}
\usepackage{amssymb}
\usepackage{float}
\usepackage{xcolor}
\newcommand*{\fullref}[1]{\hyperref[{#1}]{\autoref*{#1} \nameref*{#1}}}
\usepackage{verbatim}
\usepackage{amsmath}
\usepackage{natbib}
\usepackage{enumitem}
\usepackage{authblk}

\usepackage{amsthm}
\allowdisplaybreaks[4]
\newtheorem{theorem}{Theorem}

\newtheorem{lemma}{Lemma}
\newtheorem{corollary}{Corollary}
\newtheorem{assumption}{Assumption}
\usepackage{xcolor}

\def\*#1{\boldsymbol{#1}}

\usepackage[accepted]{icml2021}

\icmltitlerunning{Variance Reduced Training with Stratified Sampling
for Forecasting Models}
\begin{document}

\twocolumn[
\icmltitle{Variance Reduced Training with Stratified Sampling \\
for Forecasting Models}

\begin{icmlauthorlist}
\icmlauthor{Yucheng Lu}{cornell,aws}
\icmlauthor{Youngsuk Park}{aws}
\icmlauthor{Lifan Chen}{aws}
\icmlauthor{Yuyang Wang}{aws}
\icmlauthor{Christopher De Sa}{cornell}
\icmlauthor{Dean Foster}{amazon,penn}
\end{icmlauthorlist}

\icmlaffiliation{cornell}{Department of Computer Science, Cornell University, Ithaca, NY, USA.}
\icmlaffiliation{aws}{Amazon Web Services (AWS) AI Labs, Palo Alto, CA, USA.}
\icmlaffiliation{amazon}{Amazon Research, New York, NY, USA.}
\icmlaffiliation{penn}{University of Pennsylvania, Philadelphia, PA, USA}

\icmlcorrespondingauthor{Yucheng Lu}{yl2967@cornell.edu}
%\icmlcorrespondingauthor{Eee Pppp}{ep@eden.co.uk}

\icmlkeywords{variance reduction, sgd, time series, forecasting, stratified sampling}
\vskip 0.3in
]

\printAffiliationsAndNotice{This work was done during Yucheng Lu’s internship at Amazon. Lifan Chen participated in this work while he was with Amazon.}

\begin{abstract}
In large-scale time series forecasting, one often encounters the situation where the temporal patterns of time series, while drifting over time, differ from one another in the same dataset. In this paper, we provably show under such heterogeneity, training a forecasting model with commonly used stochastic optimizers (e.g. SGD) potentially suffers large variance on gradient estimation, and thus incurs long-time training. We show that this issue can be efficiently alleviated via \emph{stratification}, which allows the optimizer to sample from pre-grouped time series strata.  For better trading-off gradient variance and computation complexity, we further propose SCott (\textbf{S}tochastic Stratified \textbf{Co}ntrol Variate Gradien\textbf{t} Descen\textbf{t}), a variance reduced SGD-style optimizer that utilizes stratified sampling via control variate.  In theory, we provide the convergence guarantee of SCott on smooth non-convex objectives. Empirically, we evaluate SCott and other baseline optimizers on both synthetic and real-world time series forecasting problems, and demonstrate SCott converges faster with respect to both iterations and wall clock time.

\end{abstract}

\section{Introduction}\label{introduction}
Large-scale time series forecasting is prevalent in many real-world applications, such as traffic flow prediction \cite{vlahogianni2014short}, stock price monitoring \cite{box2011time}, weather forecasting \cite{xu2019probabilistic}, etc.
%These applications exclusively involve training a single forecasting model: assuming the future values of interests given the past observations in all the time series follows a static distribution or mapping, a \emph{single} forecasting model is expected to approximate such distribution or mapping after proper training \cite{rangapuram2018deep}.
Traditional forecasting models such as SSM \cite{durbin2012time}, ARIMA \cite{zhang2003time}, ETS \cite{de2011forecasting} and Gaussian Processes \cite{brahim2004gaussian} are the folklore methods for modeling the dynamics of a single time series.
Recently, deep forecasting models \cite{faloutsos2019forecasting} that leverage deep learning techniques have been proven to be particularly well-suited at modeling over an entire collection of time series \cite{rangapuram2018deep,wang2019deep,salinas2020deepar}. In such setting, multiple time series are jointly learned, which enables forecasting over a large scope.

In practice, a time series dataset can be heterogeneous with respect to a single forecasting model \cite{lee2018interpretable}.
The heterogeneity here specifically indicates the underlying distribution of interests may vary across different time series instances due to local effects \cite{wang2019deep, Rajat2019}; or is correlated to time in each time series individually -- a phenomenon we refer to as \emph{concept drift} \cite{gama2014survey}.
In light of this, a seemingly plausible solution is to maintain multiple forecasters. 
However, in most applications training a single model is inevitable since deploying multiple models incurs storage overhead and sometimes generalizes worse \cite{montero2020principles,oreshkin2019n,gasthaus2019probabilistic}. 
As a first investigation in this paper, we provably show the time series heterogeneity can induce arbitrarily large gradient estimation variance in many optimizers, including SGD \cite{bottou2010large}, Adam \cite{kingma2014adam}, AdaGrad \cite{ward2019adagrad}, etc.

%Aside from heterogeneity, another challenge in modern time series dataset is its large size. For instance, the \texttt{EC-full} dataset from Amazon collects the online sales of 500K different products over 300 days \cite{seeger2016bayesian}. 
%At this scale, stochastic optimizers are usually adopted in model training, which iteratively modify the model parameters towards an optimum point via repeatedly computing stochastic gradients over some predefined loss functions. 
%In the context of training deep forecasting models, a stochastic gradient generally means back propagation on a mini-batch of randomly sampled time series segments \cite{tealab2018time}. 
%Popular stochastic optimizers include SGD \cite{bottou2010large}, Adam \cite{kingma2014adam}, AdaGrad \cite{ward2019adagrad}, etc. These optimizers enjoy low computation costs and are observed to achieve better generalization in many cases \cite{zhou2020towards}.

%This happens because, as we will provably show in Section~\ref{section motivation}, the heterogeneity makes the random sampler suffer from additional noise as training examples are not close in the sense of Euclidean distance on their gradients. 

Extensive study has been conducted on reducing gradient estimation variance in stochastic optimization such as using mini-batching \cite{gower2019sgd}, control variate \cite{johnson2013accelerating} and importance sampling \cite{csiba2018importance}. These methods are mostly motivated by optimization theory and do not consider time series heterogeneity at a finer-grained level.
In this paper, we take a different perspective: observing that the distribution of interests in time series is usually recurring over time horizon or is correlated over instances \cite{liao2005clustering,aghabozorgi2015time,maharaj2019time}, we argue gradient variance induced by time series heterogeneity can be mitigated via \emph{stratification}.
Specifically, the intuition is that if we can somehow stratify the time series into multiple strata where each stratum contains homogeneous series, then the variance on the gradient estimation can be provably reduced via weighted sampling over all the strata.
Our paper concludes with a specific algorithm named SCott (\textbf{S}tochastic Stratified \textbf{Co}ntrol Variate Gradien\textbf{t} Descen\textbf{t}), an SGD-style optimizer that utilizes this stratified sampling strategy with control variate to balance variance-complexity trade off.

%To address this, we propose an alternative sampler named \emph{Subgroup Sampling},
%which samples over pre-grouped time series. 
%The intuition of subgroup sampling is that the distribution of interests in time series is usually repeating over time horizon or is correlated over instances \cite{liao2005clustering,aghabozorgi2015time,maharaj2019time}.
%With proper pre-grouping on the time series examples, the sampler is then able to capture the diversity of the underlying data more efficiently while avoiding sampling similar and redundant gradients repetitively.
%Furthermore, we propose SCott (\textbf{S}tochastic Stratified \textbf{Co}ntrol Variate Gradien\textbf{t} Descen\textbf{t}), an SGD-style optimizer that uses a control variate to utilize subgroup sampling while preserving small computation complexity.

Our contributions can be summarized as follows:
\begin{enumerate}[nosep,leftmargin=12pt]
    \item We show in theory that even on a simple AutoRegressive (AR) forecasting model, the gradient estimation variance can be arbitrarily large and slows down training.
    \item We conduct a comprehensive study on temporal time series, and show how stratification over timestamps allows us to obtain homogeneous strata with negligible computation overhead.
    \item We propose a variance-reduced optimizer SCott based on stratified sampling, and prove its convergence on smooth non-convex objectives.
    \item We empirically evaluate SCott
    on both synthetic and real-world forecasting tasks. We show SCott is able to speed up SGD, Adam and Adagrad without compromising the generalization of forecasting models.
\end{enumerate}

%%%\vspace{-10pt}
\textbf{Notations.}
Throughout this paper, we use $\*y_j$ to denote the $j$-th coordinate of a vector $\*y$. We use $\*y_{i, a:b}$ to denote $[\*y_{i,a},  \*y_{i,a+1}, \cdots, \*y_{i,b-1}, \*y_{i,b}]$. 
For two variables $g_1$ and $g_2$, $g_1=\Omega(g_2)$ means there exists a numerical constant $c$ such that $g_1\geq cg_2$. 
We use $|\mathcal{S}|$ to denote the cardinality of a set $\mathcal{S}$. We use $\mathbb{E}[X]$ and $\text{Var}[X]$ to denote the expectation and variance of a random variable $X$, given their existence.

\section{Related Work}\label{sec:related work}

\textbf{Learning from Heterogeneous Time Series.}
Heterogeneity in time series forecasting has been investigated in prior arts. Previous works mostly address this from two aspects: (1) Maintaining multiple models. A representatitive work is ESRNN \cite{smyl2020hybrid}, the M4 forecasting competition winner that proposes using ensemble of experts \cite{hewamalage2021recurrent}; 
(2) Modifying the model architecture that characterizes the time series prior to training \cite{bandara2020forecasting,chen2020probabilistic,bandara2020lstm,lara2021experimental}.
In this paper, we investigate an orthogonal direction on variance-reduced gradients. Our method does not require multiple models or modification of model architectures. 

\textbf{Sampling in Stochastic Optimization.}
In the domain of stochastic optimization, uniform sampling is the folklore sampler used in many first-order optimizers, e.g. SGD \cite{zhang2004solving}. Based on that, \citet{nagaraj2019sgd} discusses uniform sampler without replacement, \citet{gao2015active} proposes adopting an active weighted sampler for training and
\citet{park2020linear} discusses sampling with cyclic scanning. Several fairness-aware samplers are also investigated in \cite{iosifidis2019fae,wang2020towards,holstein2019improving}.
In other works, \citet{london2017pac,abernethy2020adaptive} study the effect of adaptive sampling on model generalization. A series of works extensively discuss the importance sampling based on gradient norm \cite{alain2015variance}, gradient bound \cite{lee2019meta}, loss \cite{loshchilov2015online}, etc, is able to accelerate training.
Perhaps the closest works to this paper are \cite{zhao2014accelerating,zhang2017determinantal}, which propose using stratified sampling for more diverse gradients. This, however, is notably different from our investigation as we do not use stratified sampling for mini-batching, and we focus on efficient stratification on time series data.

\textbf{Stratification in Machine Learning.}
Stratification is a powerful technique for machine learning.
For instance, application-driven works like \citet{liberty2016stratified} proposes using stratified sampling to solve a specialized regression problem in databases whereas \citet{yu2019near} discusses stratification in weakly supervised learning.
In terms of variance-reduced training, most of the previous works exclusively focus on using stratification for diversifying mini-batches. Concretely,
with the basic proposition of stratified mini-batching from \cite{zhao2014accelerating}, subsequent works like \citet{zhang2017determinantal} extends that to a sampling framework; \citet{liuaccelerating} proposes using adaptive strata; and \citet{fu2017cpsg} discusses trasferring stratification framework to Bayesian learning.

\begin{table*}[t]
\caption{Quantity of interests to approximate in different forecasting types. Inside the table, $F$ denotes the model where it takes context and features as input, and then make predictions via model parameters $\*\theta$.
$\tau_c$ and $\tau_p$ denote the context length and prediction length, $t_0$ is referred to as prediction time by convention. }
\label{table forecasting type}
\small
\begin{center}
\begin{tabular}{ccccccc}
\toprule
Forecasting Model Type & Mappings/Distributions to Approximate\\
\midrule
Deterministic & $\*{\hat{z}}_{i,t_0+1:t_0+\tau_p}=F(\*z_{i,t_0-\tau_c+1:t_0},\*x_{i,1:T_i};\*\theta)$ \\
Probabilistic & $\hat{\mathbb{P}}(\*z_{i,t_0+1:t_0+\tau_p}|\*z_{i,t_0-\tau_c+1:t_0},\*x_{i,1:T_i};\*\theta)=F(\*z_{i,t_0-\tau_c+1:t_0},\*x_{i,1:T_i};\*\theta)$ \\
\bottomrule
\end{tabular}
\end{center}
%%\vspace{-15pt}
\end{table*}
\section{Preliminaries}\label{section preliminary}
In this section we introduce the formulation of training forecasting models and stochastic gradient optimizers.

\textbf{Problem Statement.}
As in other machine learning tasks, training forecasting models is often formulated into the Empirical Risk Minimization (ERM) framework.
%Let $\mathcal{D}$ denote the set of training examples generated from 
Given $N$ different time series: $\{\*z_{i}\}_{i=1}^{N}$ where $\*z_{i,t}$ denotes the value of $i$-th time series at time $t$, let $\*x_{i,t}$ denote the (potentially) available features of $i$-th time series at time $t$, we aim to train a forecasting model $F$ with parameters $\*\theta$ (Table~\ref{table forecasting type}).
The training is then formulated by connecting $F$ with a loss function $\mathcal{L}$ to be minimized. For instance, given the notation in Table~\ref{table forecasting type}, a deterministic model over loss function $\mathcal{L}$ at prediction time $t_0$ can be expressed as 
\begin{align*}
    f_{i,t_0}(\*\theta)=\mathcal{L}(\*{z}_{i,t_0+1:t_0+\tau_p}, \hat{ \*{z}}_{i,t_0+1:t_0+\tau_p}),
\end{align*}
where $\*{\hat{z}}_{i,t_0+1:t_0+\tau_p} = F(\*z_{i,t_0-\tau_c+1:t_0},\*x_{i,1:T_i};\*\theta)$ and $f_{i,t_0}(\*\theta):\mathbb{R}^d\rightarrow \mathbb{R}$ is the loss incurred on the $i$-th time series at time $t_0$.
Popular options for loss functions $\mathcal{L}$ include Mean Square Error (MSE) Loss, Quantile Loss, Negative Log Likelihood, KL Divergence, etc \cite{gneiting2014probabilistic}. The training is then formulated as an optimization problem as
\begin{equation}\label{equation_objective}
    \*{\hat{\theta}} = \arg\min_{\*\theta\in\mathbb{R}^d}\left[f(\*\theta) = \frac{1}{|\mathcal{D}|}\sum_{i=1}^{N}\sum_{t=1}^{T_i}f_{i,t}(\*\theta)\right],
\end{equation}
where
$T_i$ denotes the maximum prediction time in the $i$-th time series. We denote $\mathcal{D}$ as the set of all the training examples indexed by $(i,t), \forall 1\leq i \leq N, \forall 1\leq t\leq T_i$.
A time series segment with certain length is then used as a single training example\footnote{In practice, this slice is of length that relates to the context and prediction length as shown in Table~\ref{table forecasting type}. A commonly used method to obtain all such segments is to use a sliding window strategy.}.

\textbf{Stochastic Gradient Optimizers.}
A stochastic gradient optimizer refers to an iterative algorithm that repeatedly updates model parameters with a uniformly sampled mini-batch of training examples.
Concretely, this stochastic gradient with mini-batch size $M$ is computed as follows:
\begin{align}\label{equation stochastic oracle}
%%\vspace{-10pt}
%\begingroup
    %\setlength{\abovedisplayskip}{1pt}
    %\setlength{\belowdisplayskip}{1pt}
    \nabla f_{\xi}(\*\theta) = \frac{1}{M}\sum\nolimits_{(i,t)\in\xi}\nabla f_{i,t}(\*\theta),
%\endgroup
%%\vspace{-10pt}
\end{align}
%%%\vspace{-10pt}
where $|\xi|=M$ and each element in $\xi$ is in the form of a tuple $(i,t)$ where
$i\sim\text{Uniform}[1, N]$ and $t\sim\text{Uniform}[1, T_i]$. Popular optimizers includes SGD, Adam, Adagrad, etc.

%It is easy to verify that Equation~(\ref{equation stochastic oracle}) is an unbiased estimation of the full gradient: $\mathbb{E}[\nabla f(\*\theta;\xi)]=\nabla f(\*\theta)$.
%Different SG optimizers utilize stochastic gradients differently. For example, Stochastic Gradient Descent (SGD) updates model directly with $\nabla f(\*\theta;\xi)$ while adaptive methods like Adam updates model with additional extrapolation \cite{kingma2014adam}.
%To proceed with our analysis in the rest of the paper, here we make a mild assumption on the SG optimizers we consider.

\iffalse
\begin{figure}[t]
    \centering
    \includegraphics[width=0.45\textwidth]{./section/Figure/chart.pdf}
    \caption{A toy time series example with temporal pattern. Consider a special case where $\tau_c=\tau_p=1$, each training example is of length 2. Time series slices labeled with same color are clustered as in the same subgroup, inside which training examples are expected to induce gradients with closer Euclidean distance.}
    \label{diagram}
\end{figure}
\fi
%%\vspace{-10pt}
\section{Motivation}\label{section motivation}
Based on the preliminaries, in this section we 
discuss our motivation that gradient variance in training forecasting models can be arbitrarily large and slow down training.
%; and (2) on a recurring time series, gradient stratification can be done efficiently with stratifying the timestamps.
%Using random sampler on a heterogeneous dataset can induce large variance.
Our motivation study is done with showing a complexity lower bound on the classic AutoRegressive (AR) model.
%We first show how gradient variance can be large during training and leads to slow training, and then we investigate the additional noise from applying commonly used random sampling on a heterogeneous dataset.
%\youngsuk{Better to say what we are going to tell: First we show that large variance can slow down in forecasting problem... Second, such large variance can be induced by heterogeneity of ...}
%\youngsuk{In this section, we show the training can depend on the variance of the stochastic gradient, which can be arbitrarily large. Then we show that the variance of the gradient under random sample strategy can be derived due to heterogeneous dataset. To investigate these two points, we adopt simple model: AutoRegression with MSE loss}

\textbf{Motivation example: AutoRegression with MSE Loss.}
Throughout this section, we consider $p$-th order AutoRegressive, or AR($p$) model, which can be expressed as follows:
\begin{equation}\label{equation_AR}
%%\vspace{-5pt}
    \*{\hat{z}}_{i,t} = \epsilon_t+\sum\nolimits_{j=1}^{p}\*\theta_{t-j}\*z_{i,t-j},
\end{equation}
where $\epsilon_t$ is a Gaussian noise term at time $t$.
We further consider adopting the Mean Squared Error (MSE) loss function, which is
\begin{equation}\label{equation mse loss}
    f_{i,t}(\*\theta) = \left(\*{\hat{z}}_{i,t} - \*z_{i,t}\right)^2.
\end{equation}
%\subsection{Large Variance can Slow Down Training}
We start from studying the variance of Stochastic Gradient (SG): $\text{Var}\left[\nabla f_{\xi}(\*\theta)\right]$. For any iterative stochastic gradient optimizer $\mathcal{A}$,
let $\*\theta^{(0)}$ and $\*\theta^{(t)}$ denote its initial parameters and its output model parameters after $t$ iterations respectively. Let $\xi^{(t)}$ denote the mini-batch sampled in its $t$-th iteration. We start from a mild assumption on $\mathcal{A}$.
\begin{assumption}\label{assumption_SG_method}
%\begingroup
%    \setlength{\abovedisplayskip}{1pt}
%    \setlength{\belowdisplayskip}{1pt}
    If 
    stochastic optimizer 
    $\mathcal{A}$ satisfies 
    $\left[\nabla f_{\xi^{(k)}}(\*\theta^{(k)})\right]_j=0$ for every $t>0$ and $k=1,\ldots,t$, then
    $[\*\theta^{(t)}]_j = [\*\theta^{(0)}]_j$ holds for any index of parameter $1\leq j\leq p$.
%\endgroup
\end{assumption}
Assumption~\ref{assumption_SG_method} is often referred to as "zero-respecting" in optimization theory \cite{carmon2019lower} and widely covers many popular optimizers (e.g. SGD, Adam, Adagrad, RMSProp, Momentum SGD, etc) under arbitrary hyperparameter settings. This
states that the SG optimizer $\mathcal{A}$ will not modify a certain parameter of the model unless a gradient updates it in the training.
With Assumption~\ref{assumption_SG_method}, we obtain the following theorem.
%\youngsuk{generally, don't need to mention 'weak' or 'strong'. Only needed if worth it technically or new assumption you make}

\begin{theorem}\label{theorem_toyexample}
%\begingroup
%    \setlength{\abovedisplayskip}{1pt}
%    \setlength{\belowdisplayskip}{1pt}
    For any AR($p$) model ($p\geq 1, p\in\mathbb{N}$) defined in Equation~(\ref{equation_AR}), there exists a time series dataset $\mathcal{D}$ with $\max_{i,t}|\*z_{i,t}|=\delta$, such that
    for any stochastic gradient optimizer $\mathcal{A}$ with any $\*\theta^{(0)}\in\mathbb{R}^p$ and hyperparameters, 
    for all $0<\epsilon<\frac{\delta^2p^{-\frac{5}{2}}}{8}$, $\mathcal{A}$ needs to compute at least
    \begin{equation}\label{equation_thm1_iteration}
    T=
        \Omega\left(\text{Var}\left[\nabla f_{\xi^{(0)}}(\*\theta^{(0)})\right]\right)
    \end{equation}
    number of stochastic gradients to find a $\*{\hat{\theta}}\in\mathbb{R}^p$ achieving $\mathbb{E}\|\nabla f(\*{\hat{\theta}})\| \leq \epsilon$. 
    Furthermore,
    \begin{equation}\label{equation_thm1_variance}
        \text{Var}\left[\nabla f_{\xi^{(0)}}(\*\theta^{(0)})\right]=\Omega\left(\delta^4p\right).
    \end{equation}
%\endgroup
\end{theorem}
Theorem~\ref{theorem_toyexample} provides important insights in two aspects. Specifically,
Equation~(\ref{equation_thm1_iteration}) shows if we wish to find a target model with small error, then the least number of stochastic gradients we need is lower bounded by the complexity of variance on the SG. And this conclusion holds for arbitrary hyperparameter scheduling (even with very small learning rate in gradient descent type optimizers).
On the other hand, Equation~(\ref{equation_thm1_variance}) reveals that the variance of SG can be arbitrarily large in theory, even with advanced transformation/preprocessing on the dataset such as magnitude scaler\footnote{With proper scaling, the magnitude $\delta$ can become a limited number, such as $1$, making $\delta^4$ bounded by a constant. However, the lower bound is still proportional to $p$, which can be arbitrarily large.} \cite{salinas2020deepar}.
As the magnitude of the dataset, or the order of AR model increases, the variance on the SG can increase to infinity in theory.

\begin{figure*}[!ht]
  \centering
  \subfigure[The Original Time Series]{
    \includegraphics[width=0.3\textwidth]{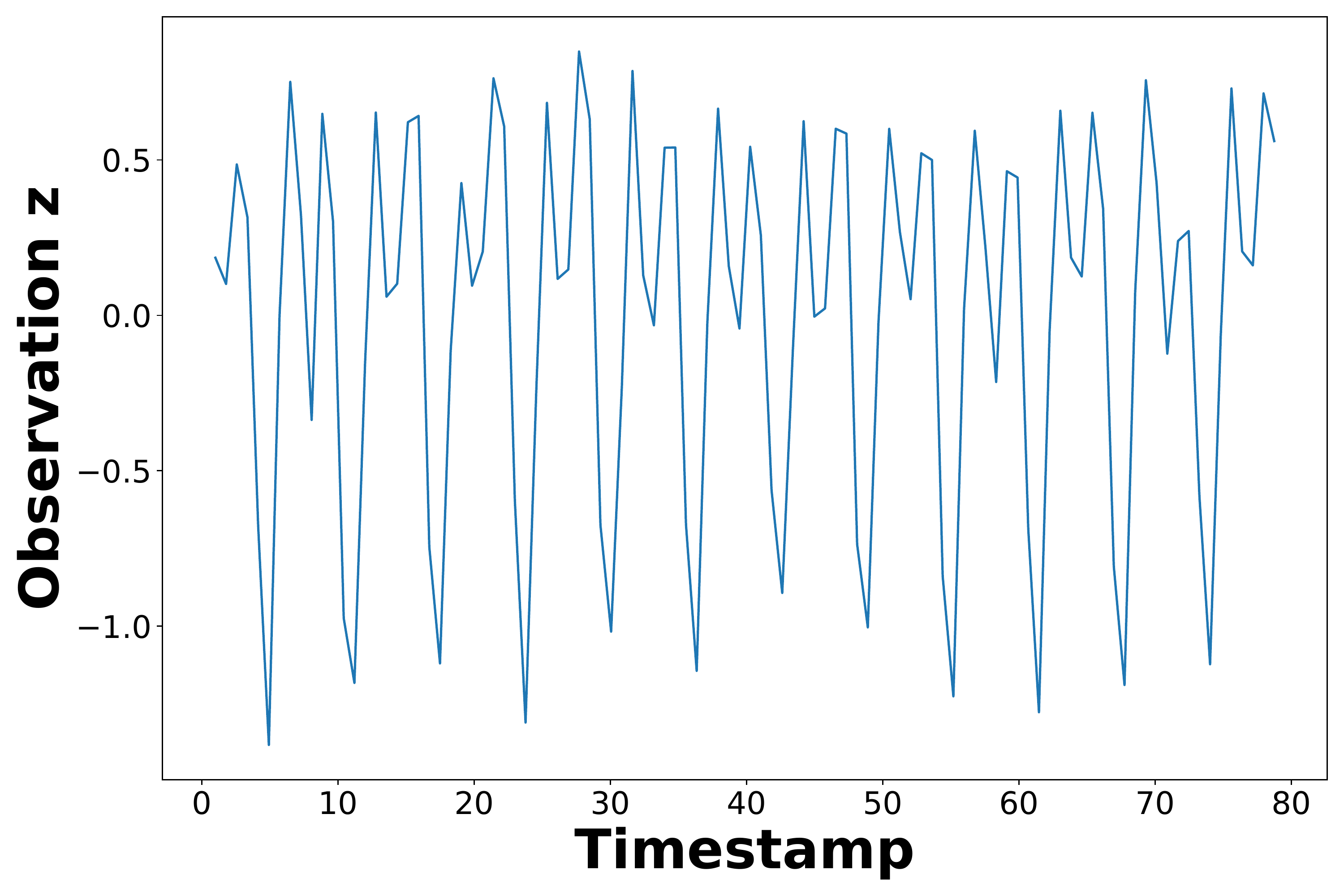}
    }
  \subfigure[Gradient Distribution]{
    \includegraphics[width=0.3\textwidth]{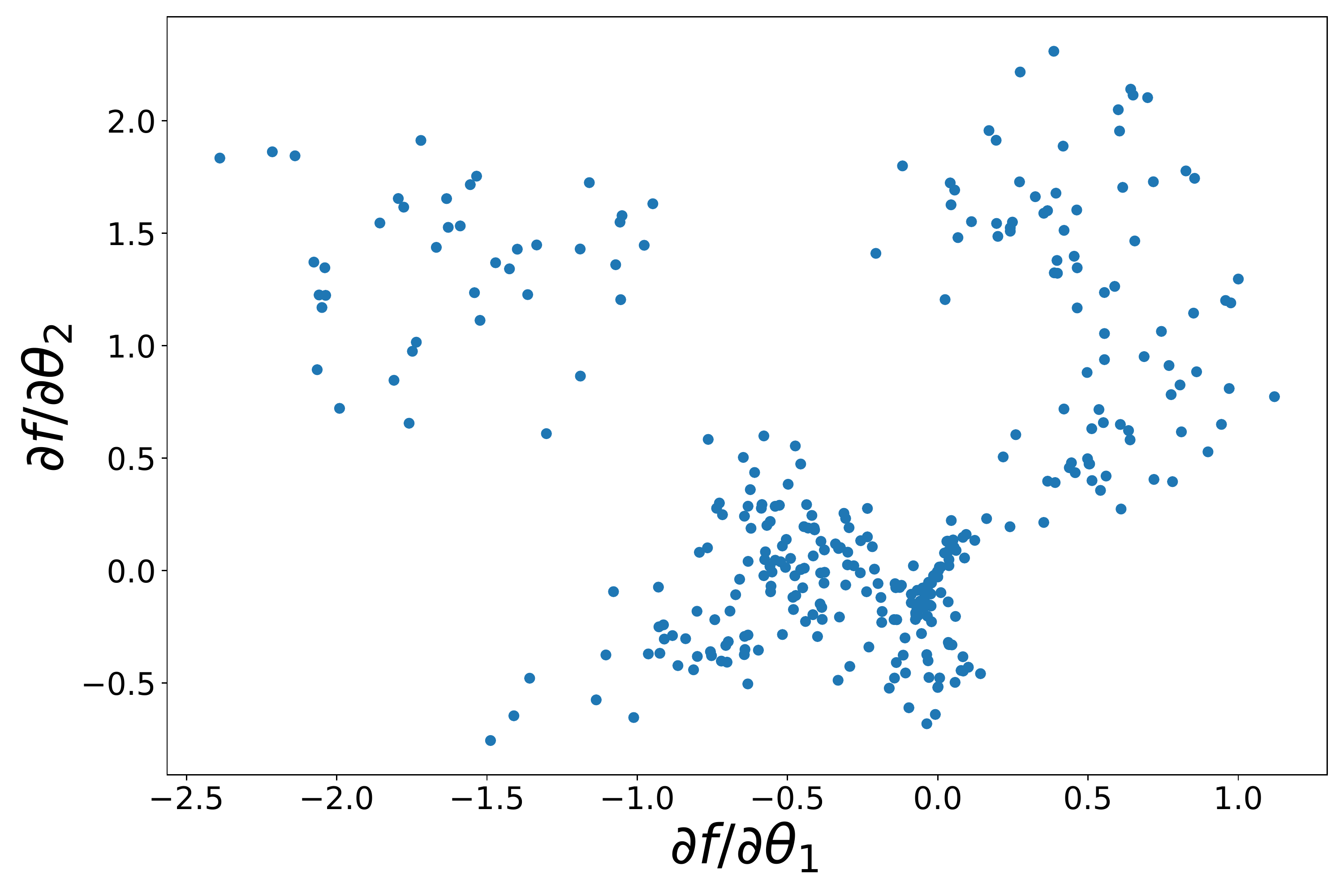}
    }
  \subfigure[Simple Stratification over Timestamps]{
    \includegraphics[width=0.3\textwidth]{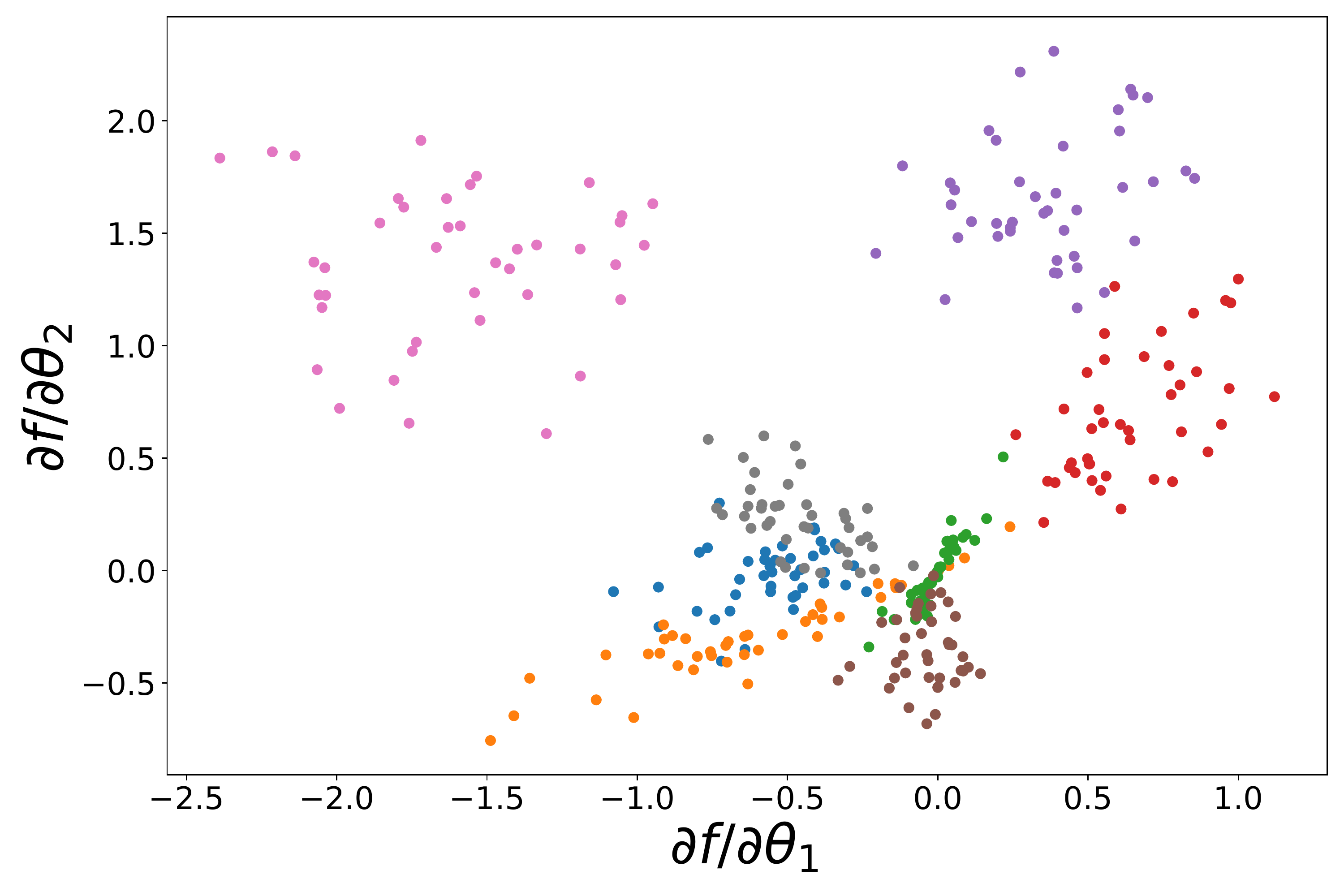}
    }
  \caption{A toy example illustrating simple stratification over the gradients via grouping the timestamps associated with each time series segment, based on estimated temporal interval in a recurring series. (Left) A time series follows the dynamics of $\sin(t) + \cos(2t) + 0.2\varepsilon_t$ with respect to time $t$ where $\varepsilon_t$ is a Gaussian noise. (Middle) The gradient distribution computed on an AR(2) model. (Right) The stratified gradients where the time series segment at time $t_0$ is grouped into the ($t_0\bmod 8$)-th stratum with $8$ strata in total.}
  \label{fig:toyexample}
\end{figure*}

\section{Training with Stratified Sampling}\label{sec:stratified_training}
Our motivation study in the previous section reveals that to provably reduce the gradient estimation variance, optimizers beyond Assumption~\ref{assumption_SG_method} should be considered. 
In this section we first illustrate comprehensively a notion of low-cost stratification over time series, and how it leads to stratified gradients. Then we introduce the stratified sampling, which provably reduces gradient variance induced from the entire heterogeneous time series instances to the inter-strata homogeneous ones.
We conclude this section with a proposition of an algorithm SCott, which modifies an existing optimizer with stratified sampling while trading-off the variance and complexity.

\subsection{Stratification over Time Series}
\textbf{Toy examples.}
While we defer the discussion on obtaining variance-reduced gradients, we start from a toy example demonstrating simple stratification over timestamps on a stationary time series instance leads to strata of the gradients.
The toy example is shown in Figure~\ref{fig:toyexample} with a simple yet effective stratification policy: hashing the timestamp based on the temporal interval. 
The intuition behind this toy example is fairly straightforward: note that in Table~\ref{table forecasting type}, the gradients induced on a certain time series segment only relates to the input features/observations, output observations and model parameters. That means, with the identical parameters, close input and output leads to close gradients in the sense of Euclidean distance.
Building on top of this insight,
if the distribution of interest is recurring over the time horizon, we can simply stratify all the time series segments based on timestamps with negligible cost.
Additionally, we provide another example in Section~\ref{section heterogeneous error} that provably illustrates the variance reduced effect through stratified sampling over timestamps.

\textbf{Generalized stratification.}
Given the toy example,
we proceed to discuss performing similar low-cost stratification over time series in general cases that leads to adequate clustering on the gradients.
A natural extension is to consider the long- and short-term temporal patterns \cite{lai2018modeling} and perform recursive stratification. For instance, if a time series instance is recurring both in terms of months and seasons, we can generalize the timestamp stratification from Figure~\ref{fig:toyexample} and perform two dimensional hashing on (months, seasons) tuples.
Note that such stratification induces much smaller overhead than clustering on high-dimensional features, since we are performing hierarchical hashing and are able to determine the stratum for each time series within logarithm time complexity. As will  be shown in the experimental section, this extended policy of stratification suffice in many settings.

On the other hand, however, finer-grained stratification can be adopted. A \emph{de facto} method is to utilize features from the dataset or extracted from the time series (e.g., density, spectrum, etc), and run a clustering algorithms such as K-Means based on that feature space \cite{bandara2020forecasting}. 
Note that based on our intuition, the main objective here is to identify time series with similar input/output as homogeneous instances while leaving the others as heterogeneous ones.
In light of this, we do not need accurate stratification in high-dimensional feature space as the previous works did \cite{aghabozorgi2015time}.

\subsection{Stratified Sampling}
Stratification outputs several strata, we next discuss how to perform low-variance gradient estimation from these strata.
Without the loss of generality, suppose the time series dataset is stratified into $B\in\mathbb{N}^{+}$ strata, i.e., 
$\mathcal{D} = \mathcal{D}_1 \cup \cdots \cup\mathcal{D}_B,$
such that each training example $(i,t)$ (recall Equation~(\ref{equation_objective})) belongs to one unique stratum.
Using the standard definition of stratified sampling, we obtain a new estimator as
\begin{align}\label{equation Stratified Sampling}
    \*{g}(\*\theta)=\sum_{i=1}^{B}\frac{|\mathcal{D}_i|\nabla f_{\xi_i}(\*\theta)}{|\mathcal{D}|},
\end{align}
where $\xi_i$ is a set of size $b$ and each element in $\xi_i$ is in the form of a tuple $(j,t)$ where $(j,t) \sim\text{Uniform }[\mathcal{D}_i]$.
Comparing Equation~(\ref{equation Stratified Sampling}) with (\ref{equation stochastic oracle}) we can see the stratified sampling essentially accumulates the examples from different strata and perform a weighted average. With simple derivation, the property of stratified sampling is summarized in the following lemma
\begin{lemma}\label{lemma Stratified Sampling}
Stratified sampler at any point $\*\theta\in\mathbb{R}^d$ satisfies
    \begin{align*}
        \mathbb{E}[\*g(\*\theta)]=\nabla f(\*\theta), \hspace{1em} \text{Var}[\*g(\*\theta)] = \sum_{i=1}^{B}\frac{|\mathcal{D}_i|^2\text{Var}[\nabla f_{\xi_i}(\*\theta)]}{|\mathcal{D}|^2}.
    \end{align*}
\end{lemma}
Lemma~\ref{lemma Stratified Sampling} reveals stratified sampling ensures the unbiased estimation of true gradient $\nabla f$, and the variance on such sampler \emph{only} depends on the variance of stochastic gradient sampled within each stratum instead of the entire dataset $\text{Var}[\nabla f_{\xi}(\*\theta)]$.
In other words, stratified sampling does not suffer additional noise even if the distribution among strata are significantly, or even adversarially different.

\begin{algorithm}[t]%\label{alg:scott}
%    \begingroup
%    \setlength{\abovedisplayskip}{1pt}
%    \setlength{\belowdisplayskip}{1pt}
	\caption{SCott (\textbf{S}tochastic Stratified \textbf{Co}ntrol Variate Gradien\textbf{t} Descen\textbf{t})}\label{svrggroup}
	\begin{algorithmic}[1]
		\REQUIRE Total number of iterations $T$, learning rate $\{\alpha_t\}_{1\leq t\leq T}$, initialized $\*\theta^{(0,0)}$, strata $\{\mathcal{D}_i\}_{1\leq i\leq B}$.
		
		%for
		\FOR{$t=0,1,\cdots, T-1$}
			\STATE  Sample a $\xi_i^{(t)}$ from stratum $i$ and perform Stratified Sampling (with $w_i=|\mathcal{D}_i|/|\mathcal{D}|$): 
			\begin{equation}\label{equation algo svrg ss}
			    \*{g}^{(t,0)}=\sum\nolimits_{i=1}^{B}w_i\nabla f_{\xi_i^{(t)}}(\*\theta^{(t,0)})
			\end{equation}
			
			%\STATE $K_t\sim$ Geom($B/(B+1)$)

			\FOR{$k=0,1,\cdots,K_t-1$}
			    \STATE Sample $\xi^{(t,k)}$ from $\mathcal{D}$.
			    \STATE Compute the update $\*v^{(t,k)}$ as
			    \begin{equation}\label{equation algo svrg inner loop}
			        \nabla f_{\xi^{(t,k)}}(\*\theta^{(t,k)})-\nabla f_{\xi^{(t,k)}}(\*\theta^{(t,0)})+\*g^{(t,0)}
			    \end{equation}
			    \STATE Update the parameters as
			    \begin{equation}\label{SCott main step}
			        \*\theta^{(t,k+1)} = \*\theta^{(t,k)} - \alpha_t\*v^{(t,k)}
			    \end{equation}
		    \ENDFOR
		    \STATE Set $\*\theta^{(t+1,0)}=\*\theta^{(t,k_t)}$.
		\ENDFOR
		\STATE \textbf{return} Sample $\hat{\*\theta}^{(T)}$ from $\{\*\theta^{(t,0)}\}_{t=0}^{T-1}$ with $\mathbb{P}(\hat{\*\theta}^{(T)}=\*\theta^{(t,0)})\propto\alpha_tB$
	\end{algorithmic}
%	\endgroup
\end{algorithm}

\subsection{SCott: Trading-off Variance and Complexity}
Despite stratified sampling mitigating the gradient variance, naively utilizing such sampler in an optimizer is suboptimal since a single sampling requires computation over $O(B\cdot M)$ gradients comparing to the $O(M)$ complexity as shown in Equation~(\ref{equation stochastic oracle}).
To address this, we propose a control variate based design on top of stratified sampling.
Our intuition is that by periodically performing a stratified sampling and computing some snapshot gradients over the training trajectory, we can use those gradients as estimation anchors to reduce variance while allowing the optimizers to adopt flexible mini-batch sizes in the effective iterations. 
In other words, we seek to achieve an intermediate solution between the plain stratified sampling and stochastic optimizers, so that we can benefit from both worlds.

The formal description of such algorithm, which we refer to as SCott, is shown in Algorithm~\ref{svrggroup}.
Note that SCott has seperate  outer and inner interation loops. A stratified sampling is only performed in each outer loop in Equation~(\ref{equation algo svrg ss}) and the output of stratified sampling $\*g^{(0,t)}$ is then used as control variate\footnote{ Refer to \cite{nelson1990control} for principles on the control variate.} in inner loop as in Equation~(\ref{equation algo svrg inner loop}).

\begin{table}[t]
\caption{Different stratification policies map SCott to algorithms SCSG \cite{li2018simple} and SVRG \cite{johnson2013accelerating} as illustrated in Section~\ref{application section}. Notations are defined in Section~\ref{application section}.}
\small
\label{table complexity}
\begin{center}
\begin{tabular}{lll}
\toprule
Stratification Policy & Complexity & Equivalence \\
\midrule
Arbitrary & Theorem~\ref{theorem scott} & SCott \\
\midrule
Random Hashing & $O\left(\Delta L\sigma^{\frac{4}{3}}\epsilon^{-\frac{10}{3}}\right)$ & SCSG \\
\midrule
Finest-Grained & $O\left(\Delta L|\mathcal{D}|^{\frac{2}{3}}\epsilon^{-2}\right)$ & SVRG \\
\bottomrule
\end{tabular}
%%\vspace{-10pt}
\end{center}
\end{table}
%%\vspace{-10pt}
\section{Convergence Analysis}\label{application section}
In this section, we 
derive the convergence rate of SCott. We first start from several assumptions.
\begin{assumption}\label{assumption smooth function}
    The loss on each single training example $f_{i,t}, \forall i, t$ is $L$-smooth:for some constant $L>0$,
    \begin{align*}
        \|\nabla f_{i,t}(\*\theta_1) - \nabla f_{i,t}(\*\theta_2)\| \leq L\|\*\theta_1 - \*\theta_2\|, \forall \*\theta_1, \*\theta_2\in\mathbb{R}^d.
    \end{align*}
\end{assumption}
Assumption~\ref{assumption smooth function} is a standard assumption in optimization theory. Note that smooth function is not necessarily convex, which implies our theory works with non-convex models, e.g. deep neural networks with Sigmoid activations.
We also make the assumption on the sampling variance as follows.
\begin{assumption}\label{assumption sampling variance}
%\begingroup
%    \setlength{\abovedisplayskip}{1pt}
%    \setlength{\belowdisplayskip}{1pt}
For all $\*\theta\in\mathbb{R}^d$, $i\in\{1,\cdots, B\}$ 
and $t=0, \cdots, T-1$ in Equation~(\ref{equation algo svrg ss}), there exists a constant $\sigma_i^2$ s.t.
    \begin{align*}
        \text{Var}_{\xi_i\sim\mathcal{D}_i}[\nabla f_{\xi_i}(\*\theta)] \leq \sigma_i^2.
    \end{align*}
%\endgroup
\end{assumption}
The constant
$\sigma_i^2$ in Assumption~\ref{assumption sampling variance} denotes the upper bound on the gradient variance when uniform sampling is performed inside the $i$-th stratum. 
%The subscripts $B$ here emphasizes the variance bound depends on the number of strata $B$.
%\youngsuk{many wrong and confusing notations. $B=|G|$ according to stratum according to SS paragraph.}
%\begin{assumption}\label{assumption variance compare}
%    For $B_1, B_2 > 0$, If $B_1\leq B_2$, then $\sum_{i=1}^{B_1}w_i^2\sigma_{B_1}^2 \geq \sum_{i=1}^{B_2}w_i^2\sigma_{B_2}^2$.
%\end{assumption}
%Assumption~\ref{assumption variance compare} shows that between two clustering policy, a finer-grained stratum will ensure a smaller inner variance. 
%Two extreme cases are: with finest-grained clustering, each stratum contains only one training example and thus the inner variance becomes zero. On the other hand, with no clustering, the variance is maximized.
For the convenience of later discussion, \textit{we further denote $\sigma^2$ as the upper bound on the variance when uniform sampling over the entire dataset}: $\text{Var}_{\xi\sim\mathcal{D}}[\nabla f_{\xi}(\*\theta)] \leq \sigma^2$.
Without the loss of generality, we let $M=1$ in our theory.
Based on the two assumptions, the convergence rate of Algorithm~\ref{svrggroup} is shown in the following theorem:
\begin{theorem}\label{theorem scott}
    Denote $\Delta=f(\*0)-\inf_{\*\theta}f(\*\theta)$ and $w_i=|\mathcal{D}_i|/|\mathcal{D}|$. 
    For any $\epsilon>0$, if we stratify the dataset $\mathcal{D}$ into $\{\mathcal{D}_i\}_{i=1}^B$ such that $\sum_{i=1}^{B}w_i^2\sigma_{i}^{2}= O\left(\epsilon^2\right)$, and let inner loop iterations $K_t \sim \text{Geo}(B/(B+1)$)\footnote{$\text{Geo}(p)$ is the geometric distribution with $p$, i.e., $\mathbb{P}(K_t=K) = p^K(1-p), \forall K=0, 1, \cdots$},
    Algorithm~\ref{svrggroup} needs to compute at most 
    \begin{align*}
    \small
        T = O\left(\frac{\Delta L\left(B\sum_{i=1}^{B}w_i^2\sigma_i^2\right)^{\frac{2}{3}}}{\epsilon^{\frac{10}{3}}}+\frac{\Delta L|\mathcal{D}|^{\frac{2}{3}}\mathbb{I}\{B=|\mathcal{D}|\}}{\epsilon^2}\right)
    \end{align*}
    number of stochastic gradients to ensure $\mathbb{E}\left\|\nabla f\left(\hat{\*\theta}^{(T)}\right)\right\|\leq\epsilon$, where $\mathbb{I}\{\cdot\}$ denotes the Indicator function.
\end{theorem}
If we deliberately let all the strata maintain the same size, the convergence rate can be simplified as follows.
\begin{corollary}\label{corollary scott}
\begingroup
    \setlength{\abovedisplayskip}{1pt}
    \setlength{\belowdisplayskip}{1pt}
    Following Theorem~\ref{theorem scott}, if all the strata are the same size, i.e.,  $|\mathcal{D}_i|=|\mathcal{D}_j|>1, \forall i, j$, Algorithm~\ref{svrggroup} needs to compute at most 
    \begin{align}
    \small
        T = O\left(\frac{\Delta L\left(B^{-1}\sum_{i=1}^{B}\sigma_i^2\right)^{\frac{2}{3}}}{\epsilon^{\frac{10}{3}}}\right)
    \end{align}
    number of stochastic gradients to ensure output $\hat{\*\theta}^{(T)}$ fulfills $\mathbb{E}\left\|\nabla f\left(\hat{\*\theta}^{(T)}\right)\right\|\leq\epsilon$.
\endgroup
\end{corollary}
%\youngsuk{compare with table}
%\youngsuk{title of remark 1,2}
%\youngsuk{Capitalization on each work in remark titles}
\textbf{Remark 1: Consistent with Other Algorithms.}
%\youngsuk{consistent in temrs of algorithm? or convergence rate? unclear}
Theorem~\ref{theorem scott} demonstrates SCott can be seen as a general form of other control variate type algorithms. Specifically, when we stratify the time series by random hashing, i.e., cyclically assign each example into $B$ strata, then SCott matches with SCSG \cite{li2018simple}. On the other hand, if we adopt finest-grained stratification, i.e., $B=|\mathcal{D}|$, then SCott will be aligned with SVRG \cite{johnson2013accelerating}.

\textbf{Remark 2: Reduced Variance Dependency.}
%Note that since $B\leq|\mathcal{D}|$ and $\sigma_{|\mathcal{D}|}^2=0$, strata with $\sum_{i=1}^{B}w_i^2\sigma_{i}^{2}= O\left(\epsilon^2\right)$ are always guaranteed to exist, as we can at least choose $B=|\mathcal{D}|$.
%\addressed{Inserted Youngsuk's opinion.}
Note that $\sum_{i=1}^{B}w_i^2\sigma_{i}^{2}= O\left(\epsilon^2\right)$ can always be fulfilled since we can at least select $B=|\mathcal{D}|$ and obtain $\sigma_{i}^2=0$, as in that case every stratum only contains one sample. 
If this precondition is somehow violated, it may only guarantee suboptimality in theory, converging to a noisy ball with $\sum_{i=1}^{B}w_i^2\sigma_{i}^{2}$. However, comparing to other stochastic control variate type optimizers, including \cite{li2018simple} and  \cite{babanezhad2015stop}, where noise ball is in the order of $O(\sigma^2)$, SCott is able to reduce the dependency only on the inner stratum variance, i.e., from $\sigma^2$ to $B\sum_{i=1}^{B}w_i^2\sigma_i^2$ (and $B^{-1}\sum_{i=1}^{B}\sigma_i^2$ with Corollary~\ref{corollary scott}).

\textbf{Remark 3: Understanding the Selection of $K_t$.} 
The number of inner loops per outer loop ($K_t$), i.e., the frequency of performing a stratified sampling is a crucial design choice. 
Theorem~\ref{theorem scott} show that a Geometric distributed selection helps with the convergence, which is a technique used in other analysis of control variate type algorithms \cite{li2018simple,horvath2020adaptivity}.
In practice, we can optimize such selection via an additional hyperparameter: in the supplementary material, we discuss using $\|\*v^{(t,k)}\|^2 \leq \gamma \|\*v^{(t,0)}\|^2$ as an additional criterion to terminate the inner loop for some hyperparameter $\gamma$.

\begin{table*}[ht]
\small
\caption{Performance with different algorithms given the same time budget for training. We compute the loss over the entire training and test dataset. We assign time budget of 0.5 and 3 hours for each setting, respectively, and present the mean and standard deviation among 5 different runs. All the optimizers are fine tuned in each task.}
\label{table experiment}
\small
\begin{center}
\begin{tabular}{clcc|cc|cccc}
\toprule
\multirow{2}{*}{Setting} & \multirow{2}{*}{Optimizer} & \multicolumn{2}{c|}{Exchange Rate} & \multicolumn{2}{c|}{Traffic} & \multicolumn{2}{c}{Electricity} \\
\cmidrule{3-8}
& & Training & Test & Training & Test & Training & Test \\
\midrule
%%%%%%%%%%%%%%%%%%%%%%
% mlp line SGD
\multirow{7}{*}{\shortstack{MLP \\ NLL}} 
& SGD & -1.825 $\pm$ 0.013 & -1.715 $\pm$ 0.017 & -2.387 $\pm$ 0.015 & -2.443 $\pm$ 0.021 & 6.512 $\pm$ 0.028 & 5.558 $\pm$ 0.021 \\
% mlp line SCSG
& SCSG & -2.037 $\pm$ 0.009 & \textbf{-1.732} $\pm$ 0.019 & -2.612 $\pm$ 0.013 & -2.674 $\pm$ 0.024 & 6.424 $\pm$ 0.017 & 5.477 $\pm$ 0.022 \\
% mlp line SCott
& SCott & \textbf{-2.145 $\pm$ 0.008} & -1.685 $\pm$ 0.009 & \textbf{-2.867 $\pm$ 0.019} &  \textbf{-2.812 $\pm$ 0.020} & \textbf{6.303 $\pm$ 0.018} & \textbf{5.348 $\pm$ 0.016} \\
\cmidrule{2-8}
% mlp line Adam
& Adam & -2.762 $\pm$ 0.009 & -2.945 $\pm$ 0.012 & -2.597 $\pm$ 0.021 & -2.593 $\pm$ 0.015 & 6.388 $\pm$ 0.025 & 5.483 $\pm$ 0.021 \\
% mlp line S-Adam
& S-Adam & \textbf{-3.917 $\pm$ 0.009} & \textbf{-3.032 $\pm$ 0.006} & \textbf{-3.038 $\pm$ 0.011} & \textbf{-3.012 $\pm$ 0.018} & \textbf{5.737 $\pm$ 0.023} & \textbf{5.015 $\pm$ 0.014} \\
\cmidrule{2-8}
% mlp line Adagrad
& Adagrad & -3.488 $\pm$ 0.011 & -3.218 $\pm$ 0.011 & -2.692 $\pm$ 0.012 & -2.763 $\pm$ 0.021 & 6.185 $\pm$ 0.019 & 5.295 $\pm$ 0.017 \\
% mlp line S-Adagrad
& S-Adagrad & \textbf{-3.886 $\pm$ 0.007} & \textbf{-3.239 $\pm$ 0.012} & \textbf{-2.864 $\pm$ 0.013} & \textbf{-2.895 $\pm$ 0.018} & \textbf{6.022 $\pm$ 0.020} & \textbf{5.182 $\pm$ 0.029} \\
\midrule
%%%%%%%%%%%%%%%%%%%%%%
% nbeats line SGD
\multirow{7}{*}{\shortstack{N-BEATS \\ MAPE}} & SGD & 1.224 $\pm$ 0.018 & 1.346 $\pm$ 0.021 & 2.798 $\pm$ 0.008 & 3.024 $\pm$ 0.019 & 0.628 $\pm$ 0.023 & 0.676 $\pm$ 0.030 \\
% nbeats line SCSG
& SCSG & \textbf{1.034 $\pm$ 0.016} & \textbf{1.182 $\pm$ 0.019} & 2.024 $\pm$ 0.012 & 2.644 $\pm$ 0.017 & 0.610 $\pm$ 0.021 & 0.635 $\pm$ 0.029 \\
% nbeats line SCott
& SCott & 1.077 $\pm$ 0.022 & 1.222 $\pm$ 0.012 & \textbf{1.898 $\pm$ 0.013} &  \textbf{2.373 $\pm$ 0.011} & \textbf{0.516 $\pm$ 0.020} & \textbf{0.546 $\pm$ 0.021} \\
\cmidrule{2-8}
% nbeats line Adam
& Adam & 0.695 $\pm$ 0.021 & 0.773 $\pm$ 0.018 & 1.013 $\pm$ 0.012 & 1.036 $\pm$ 0.019 & 0.589 $\pm$ 0.017 & 0.697 $\pm$ 0.021 \\
% nbeats line S-Adam
& S-Adam & \textbf{0.514 $\pm$ 0.012} & \textbf{0.593 $\pm$ 0.021} & \textbf{0.809 $\pm$ 0.021} & \textbf{0.813 $\pm$ 0.021} & \textbf{0.445 $\pm$ 0.025} & \textbf{0.528 $\pm$ 0.011} \\
\cmidrule{2-8}
% nbeats line Adagrad
& Adagrad & 0.764 $\pm$ 0.022 & 0.806 $\pm$ 0.012 & 2.068 $\pm$ 0.012 & 1.997 $\pm$ 0.018 & 0.537 $\pm$ 0.011 & 0.631 $\pm$ 0.025 \\
% nbeats line S-Adagrad
& S-Adagrad & \textbf{0.563 $\pm$ 0.013} & \textbf{0.692 $\pm$ 0.009} & \textbf{1.486 $\pm$ 0.014} & \textbf{1.724 $\pm$ 0.018} & \textbf{0.428 $\pm$ 0.020} & \textbf{0.512 $\pm$ 0.016} \\
\bottomrule
\end{tabular}
\end{center}
%\vspace{-15pt}
\end{table*}

\textbf{Remark 4: Improved Complexity Compared to Stochastic Optimizers.}
\citet{carmon2019lower} shows the theoretical lower bound on the complexity of stochastic optimizers is $\Omega\left(\Delta L\sigma^2\epsilon^{-4}\right)$,
Comparing Corollary~\ref{corollary scott} with this bound, we can observe a complexity improvement of at least $O\left(\epsilon^{-\frac{2}{3}}\right)$ from SCott compared to stochastic optimizers. On the other hand, as also shown in Table~\ref{table complexity}, the convergence rate of SCott improves upon SCSG and stochastic optimizers in the sense that its upper bound depends only on the inner strata variance.

%%%%%%%%%%Synthetic Dataset%%%%%%%%%%%%%%%%%%
\begin{figure*}[t]
%\centering
    \subfigure[Loss w.r.t. \# Iterations \newline VAR - Synthetic Dataset]{
        \includegraphics[width=0.23\textwidth]{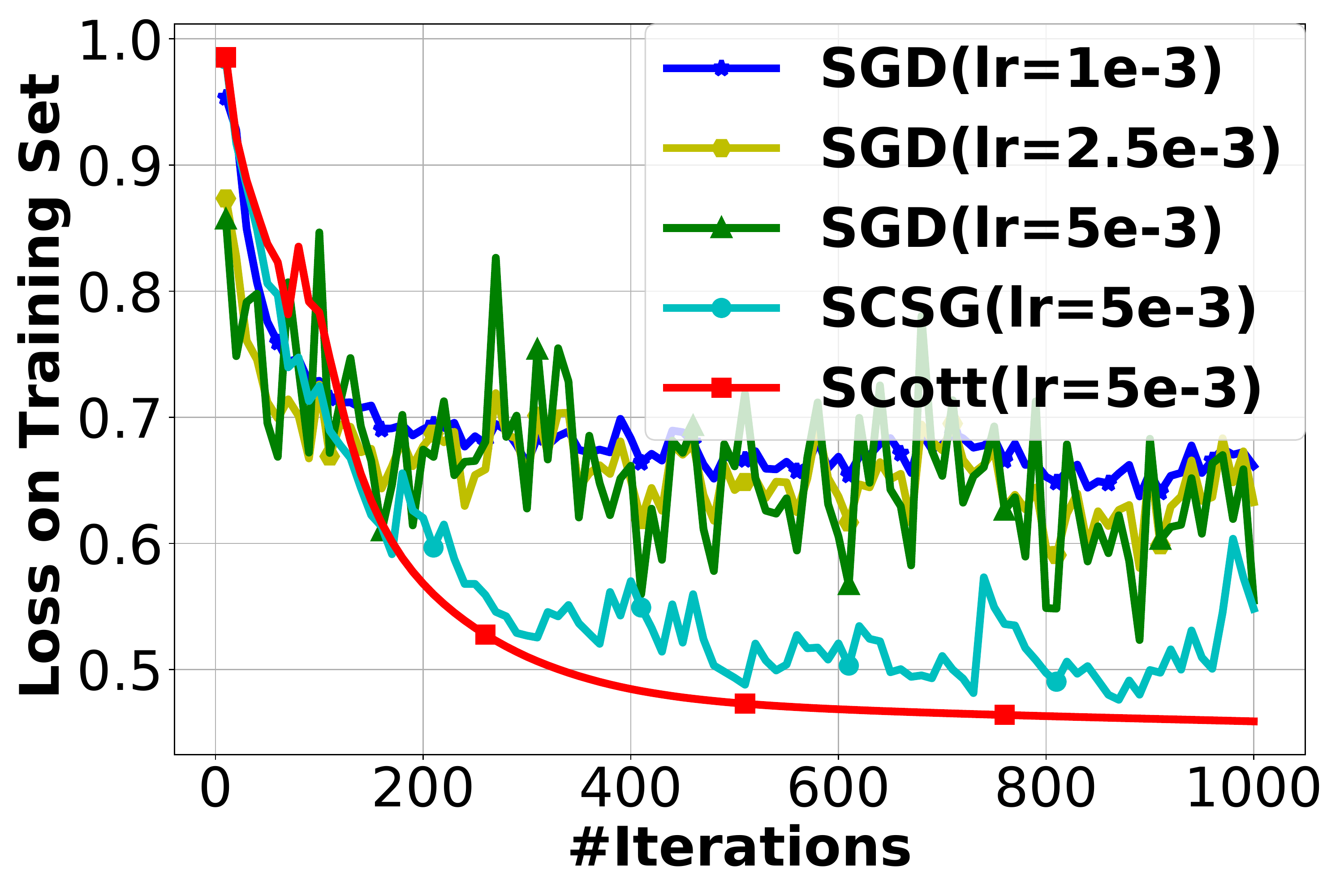}
        \label{VAR_synthetic_iteration}
    }
%\centering
    \subfigure[Loss w.r.t. Time(s) \newline VAR - Synthetic Dataset]{
        \includegraphics[width=0.23\textwidth]{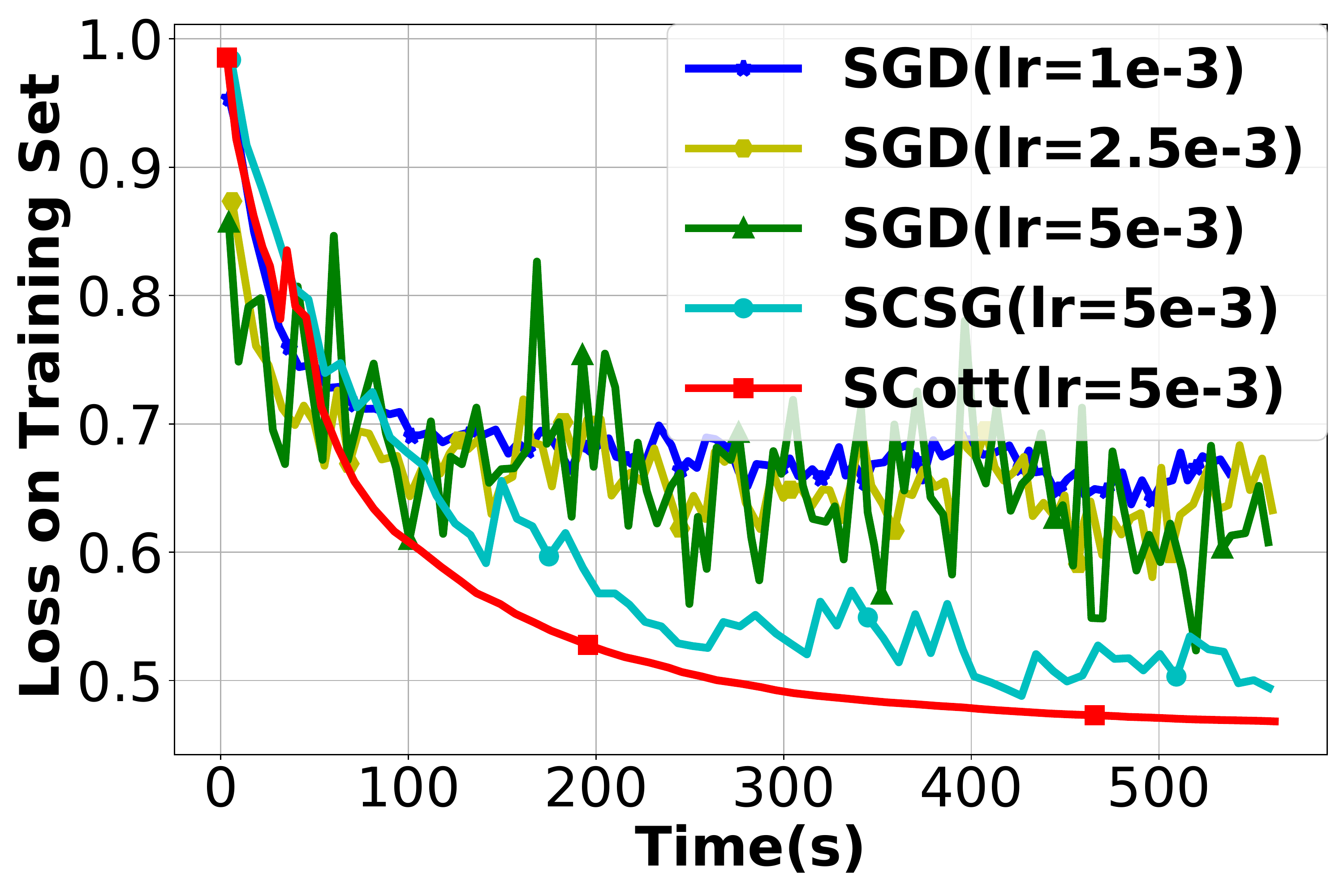}
        \label{VAR_synthetic_time}
    }
%\centering
    \subfigure[Loss w.r.t. \# Iterations \newline LSTM - Synthetic Dataset]{
        \includegraphics[width=0.23\textwidth]{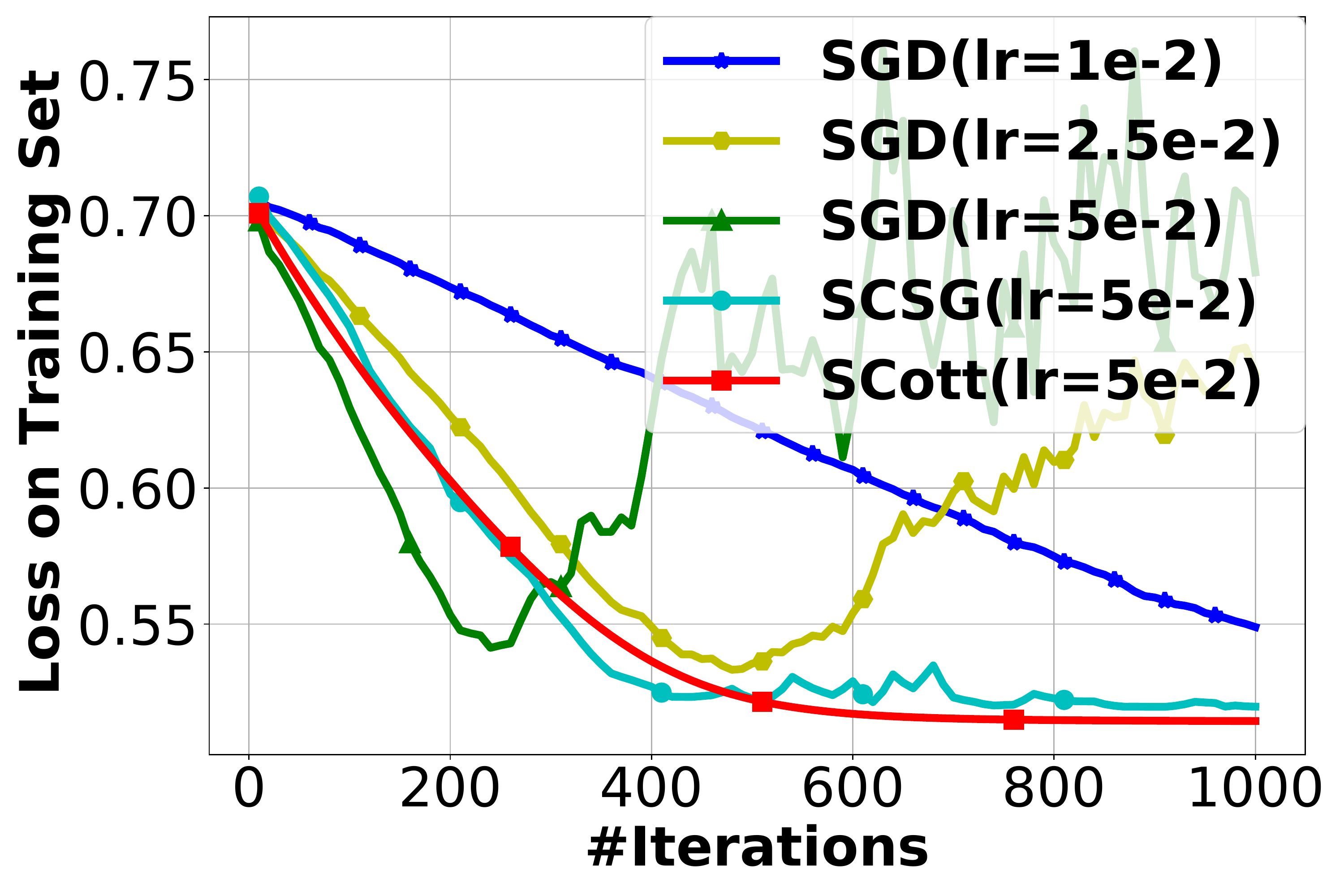}
        \label{LSTM_synthetic_iteration}
    }
%\centering
    \subfigure[Loss w.r.t. Time(s) \newline LSTM - Synthetic Dataset]{
        \includegraphics[width=0.23\textwidth]{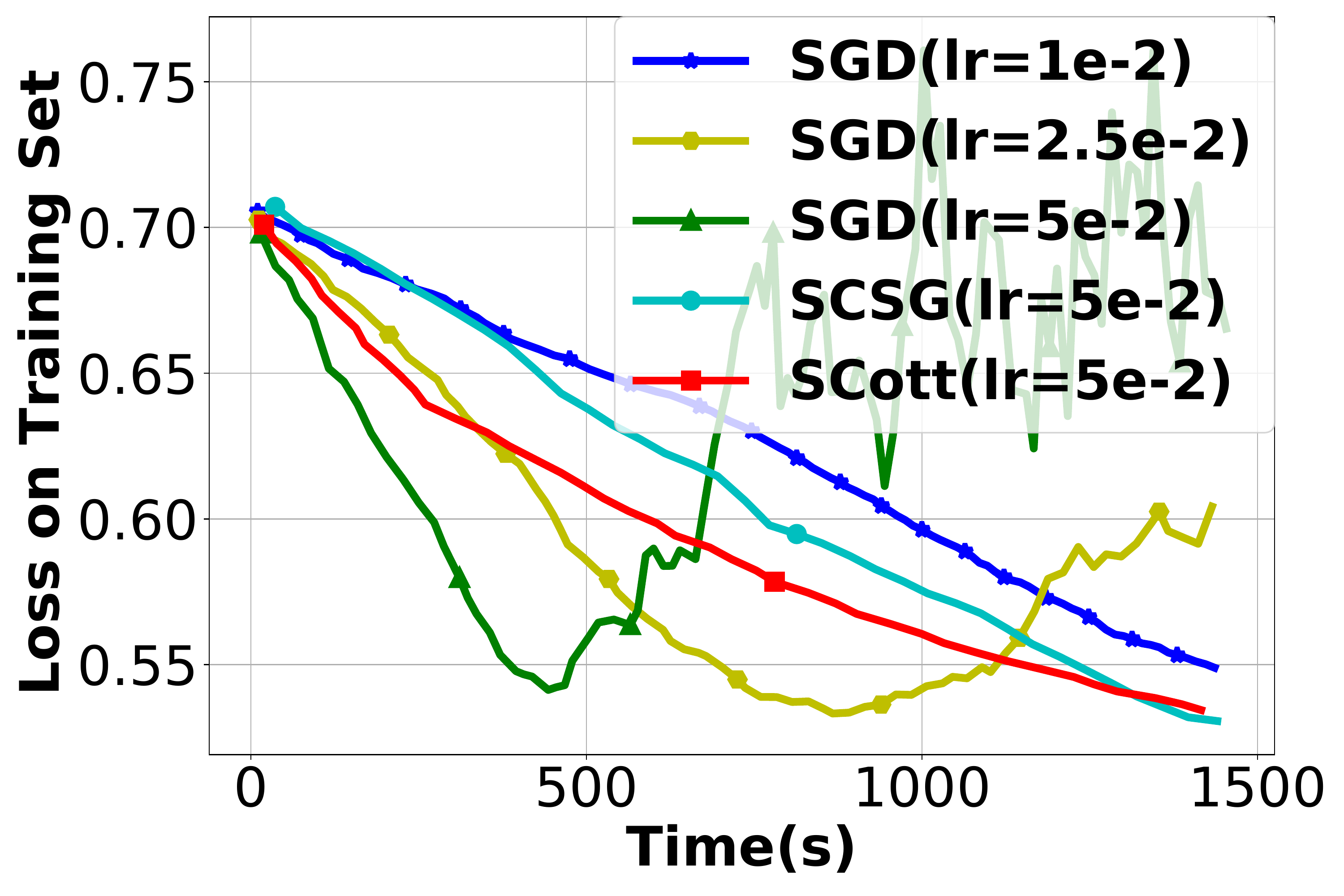}
        \label{LSTM_synthetic_time}
    }
    \subfigure[Training Loss / \# Iterations \newline MLP - Traffic Dataset]{
        \includegraphics[width=0.23\textwidth]{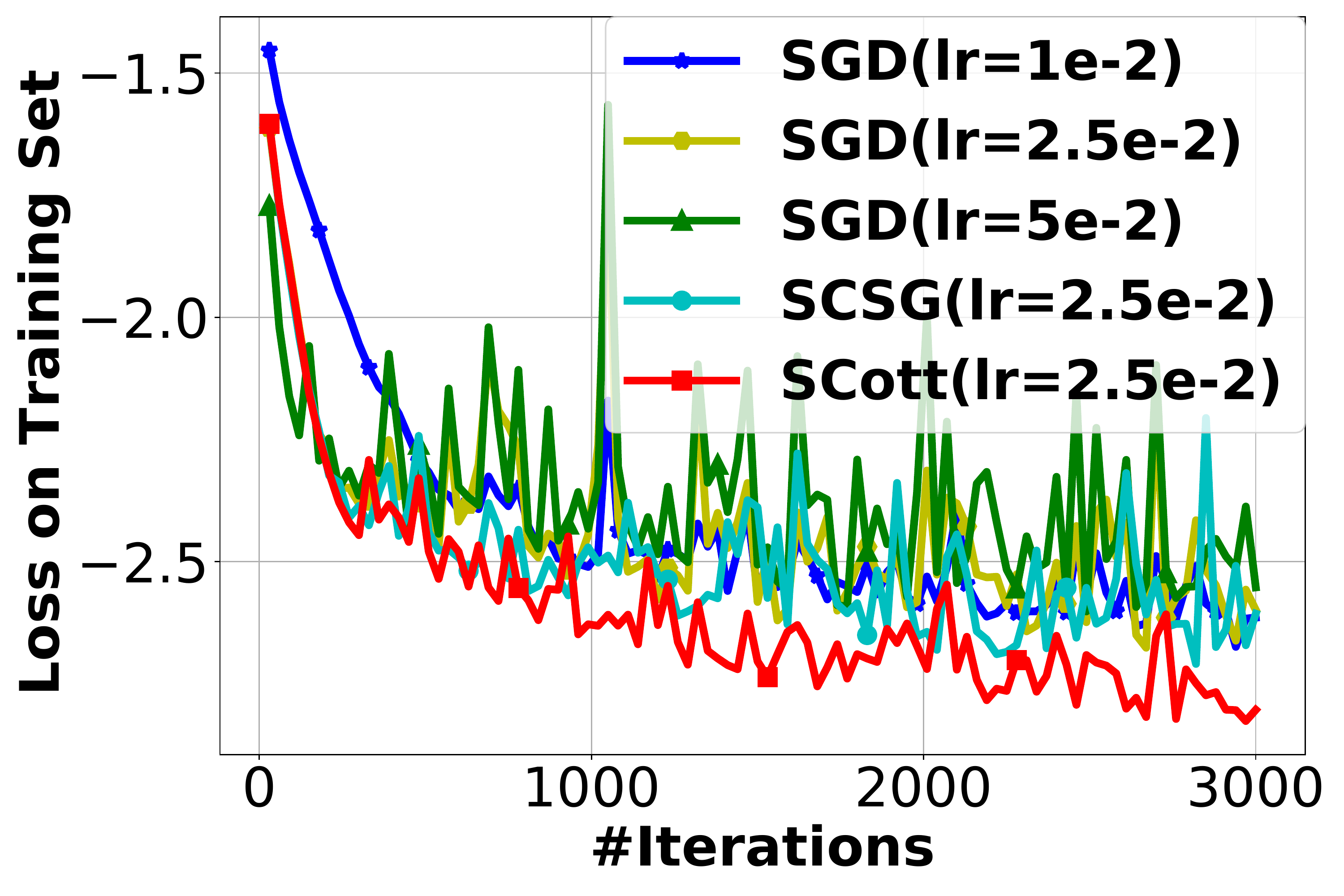}
        \label{MLP_traffic_iteration}
    }
     \subfigure[Training Loss / Time(s) \newline MLP - Traffic Dataset]{
        \includegraphics[width=0.23\textwidth]{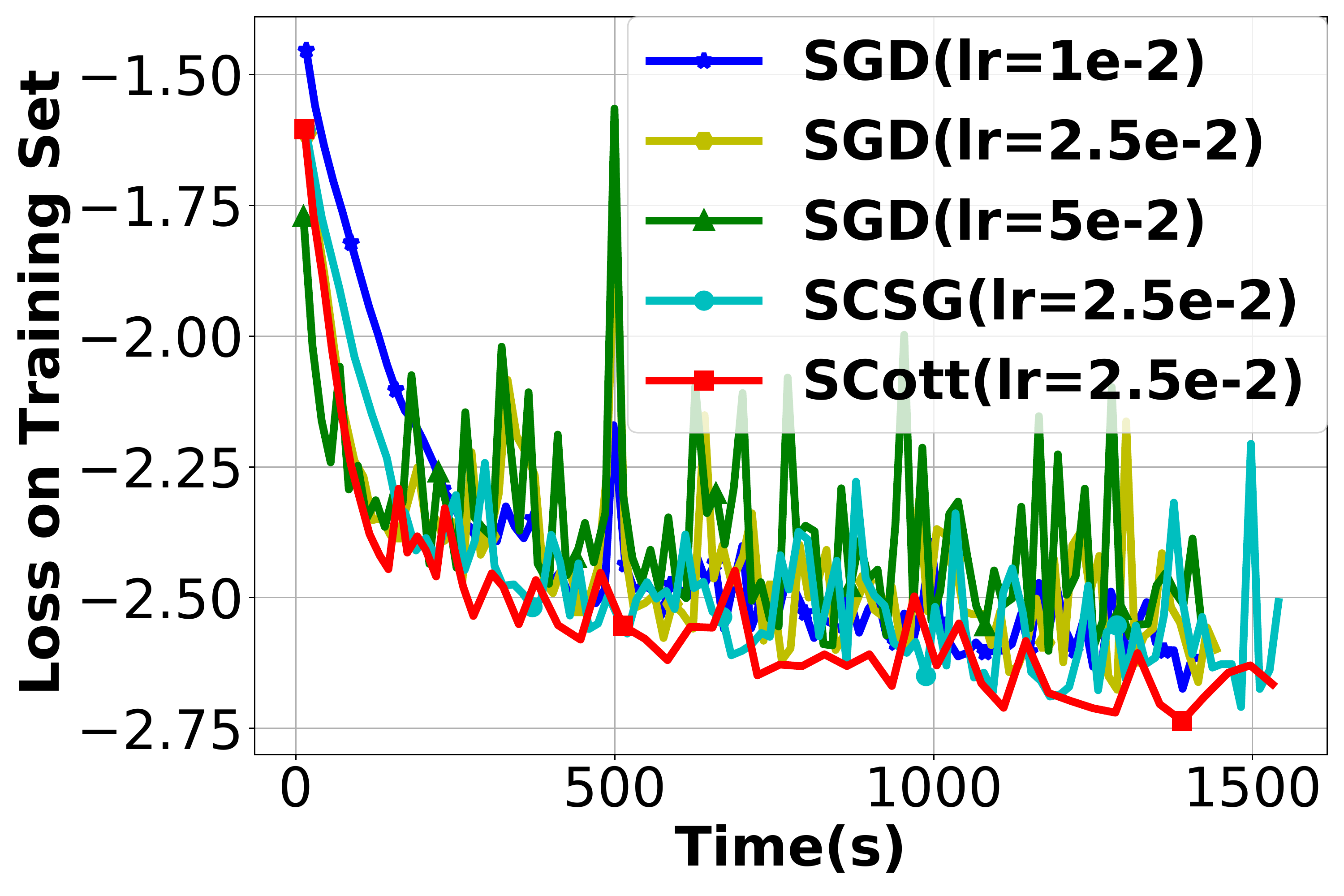}
        \label{MLP_traffic_time}
    }
%\centering
    \subfigure[Test Loss / \# Iterations \newline MLP - Traffic Dataset]{
        \includegraphics[width=0.23\textwidth]{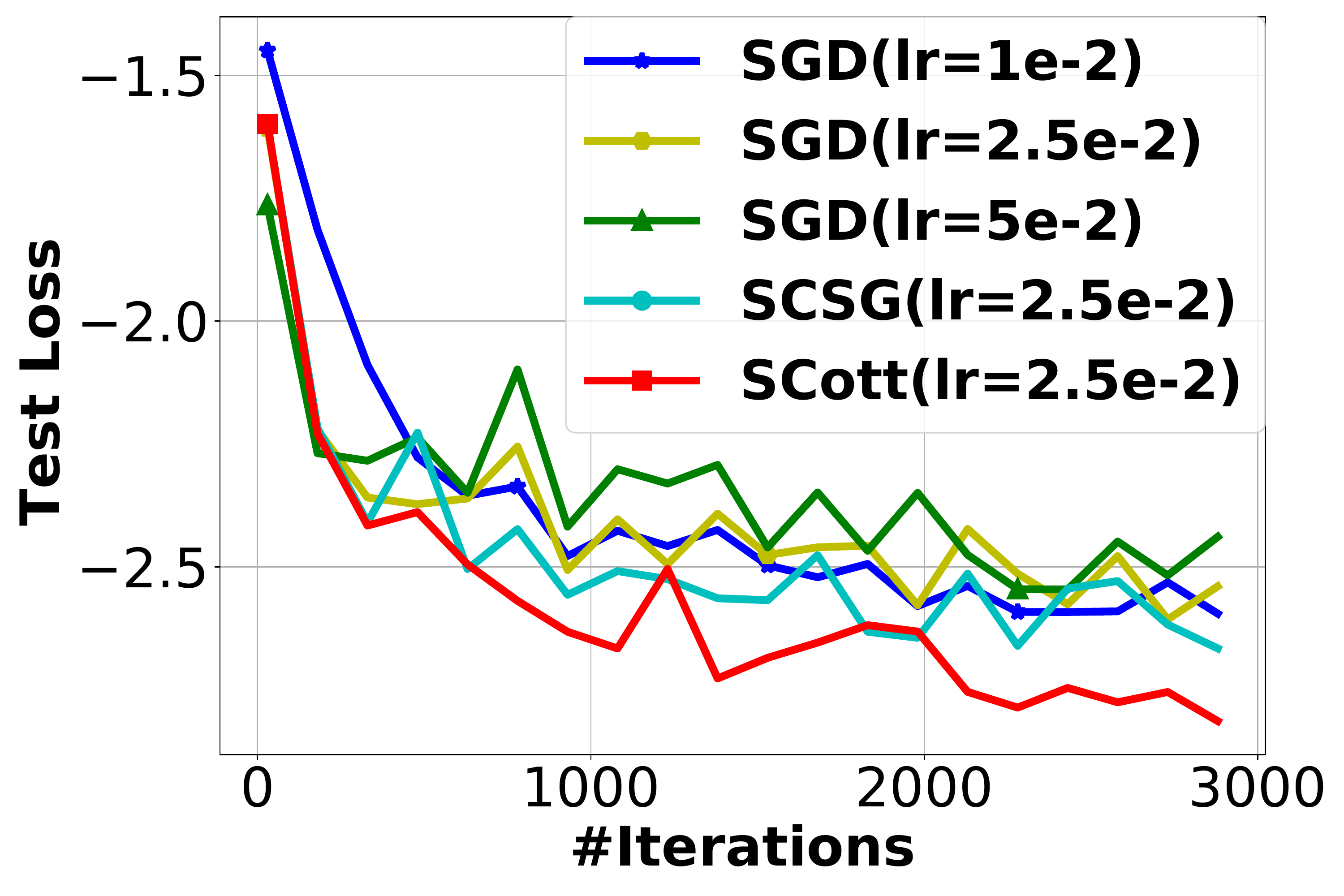}
        \label{MLP_traffic_val}
    }
     \subfigure[Variance of Samplers \newline MLP - Traffic Dataset]{
        \includegraphics[width=0.23\textwidth]{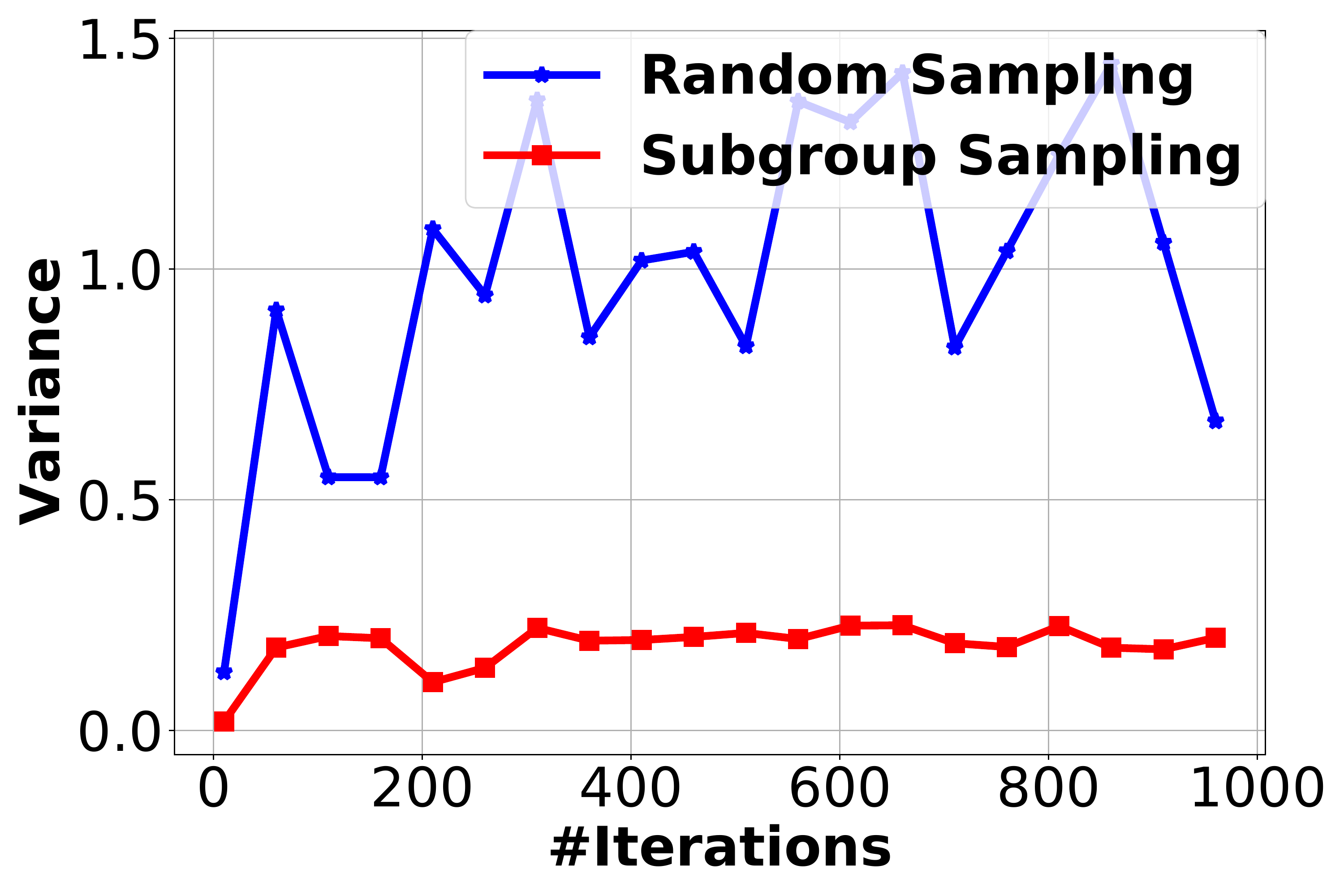}
        \label{MLP_traffic_var}
    }
    \subfigure[Training Loss / \# Iterations \newline N-BEATS - Electricity Dataset]{
        \includegraphics[width=0.23\textwidth]{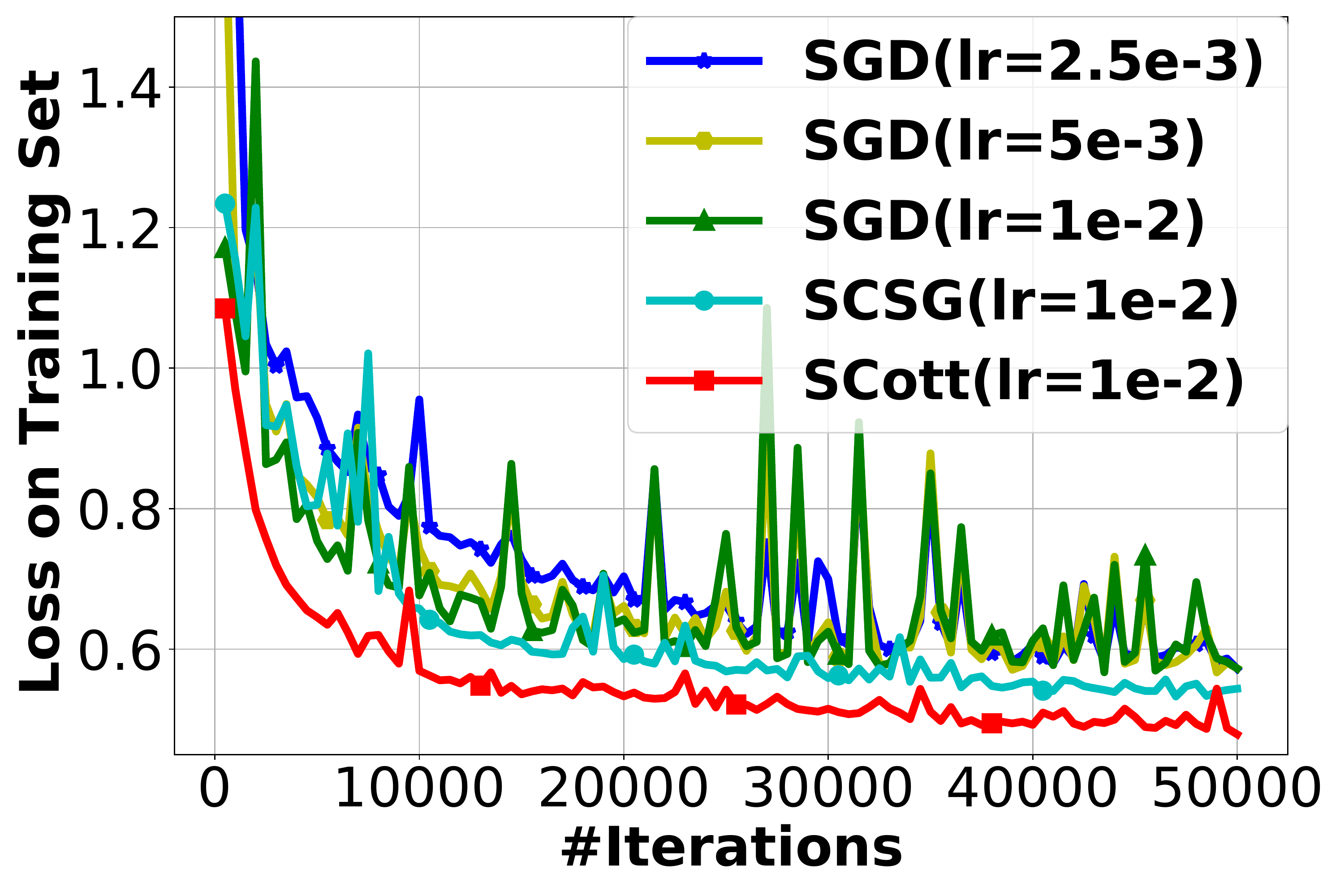}
        \label{nbeats_electricity_iteration}
    }
     \subfigure[Training Loss / Time(min) \newline N-BEATS - Electricity Dataset]{
        \includegraphics[width=0.23\textwidth]{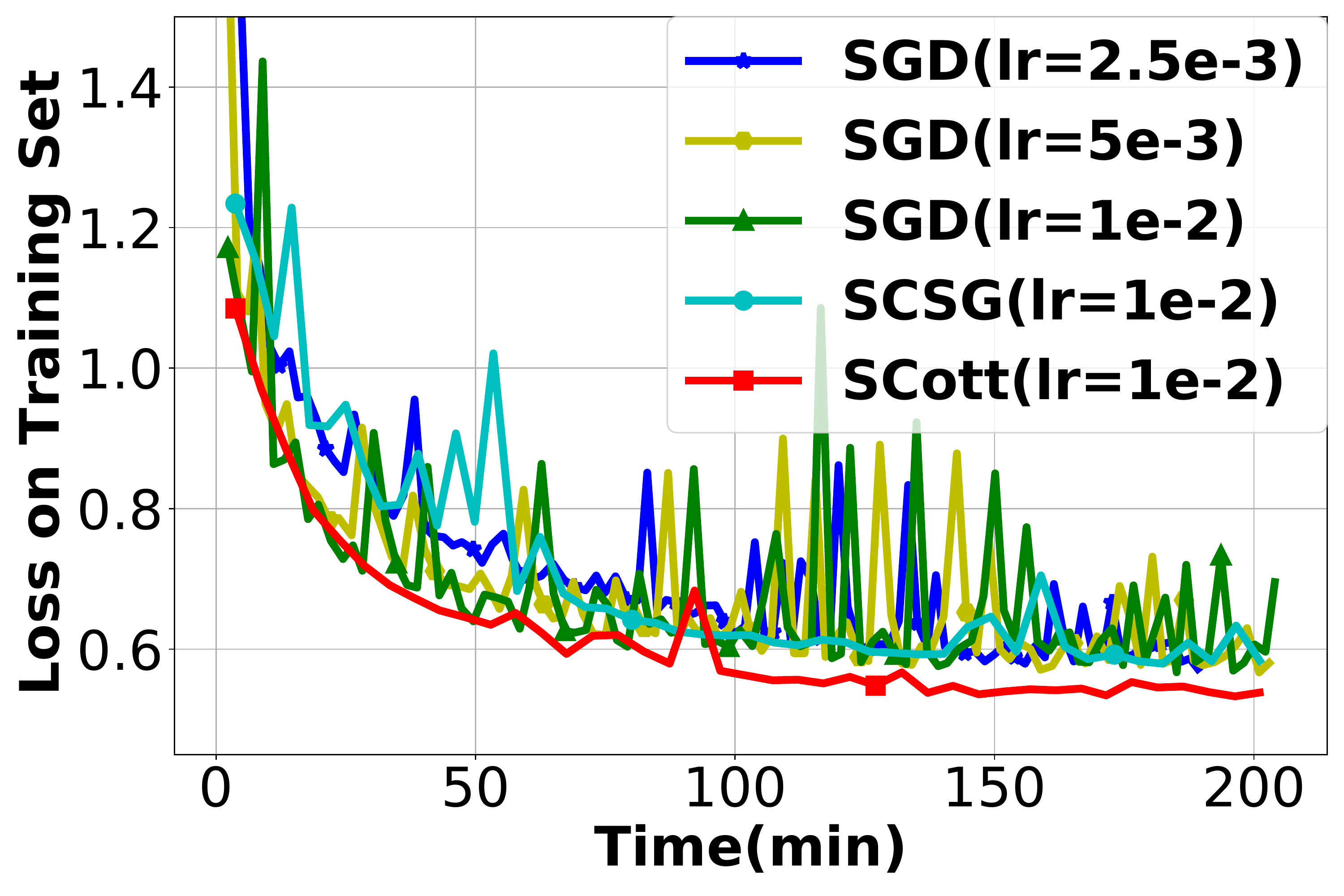}
        \label{nbeats_electricity_time}
    }
%\centering
    \subfigure[Test Loss / \# Iterations \newline N-BEATS - Electricity Dataset]{
        \includegraphics[width=0.23\textwidth]{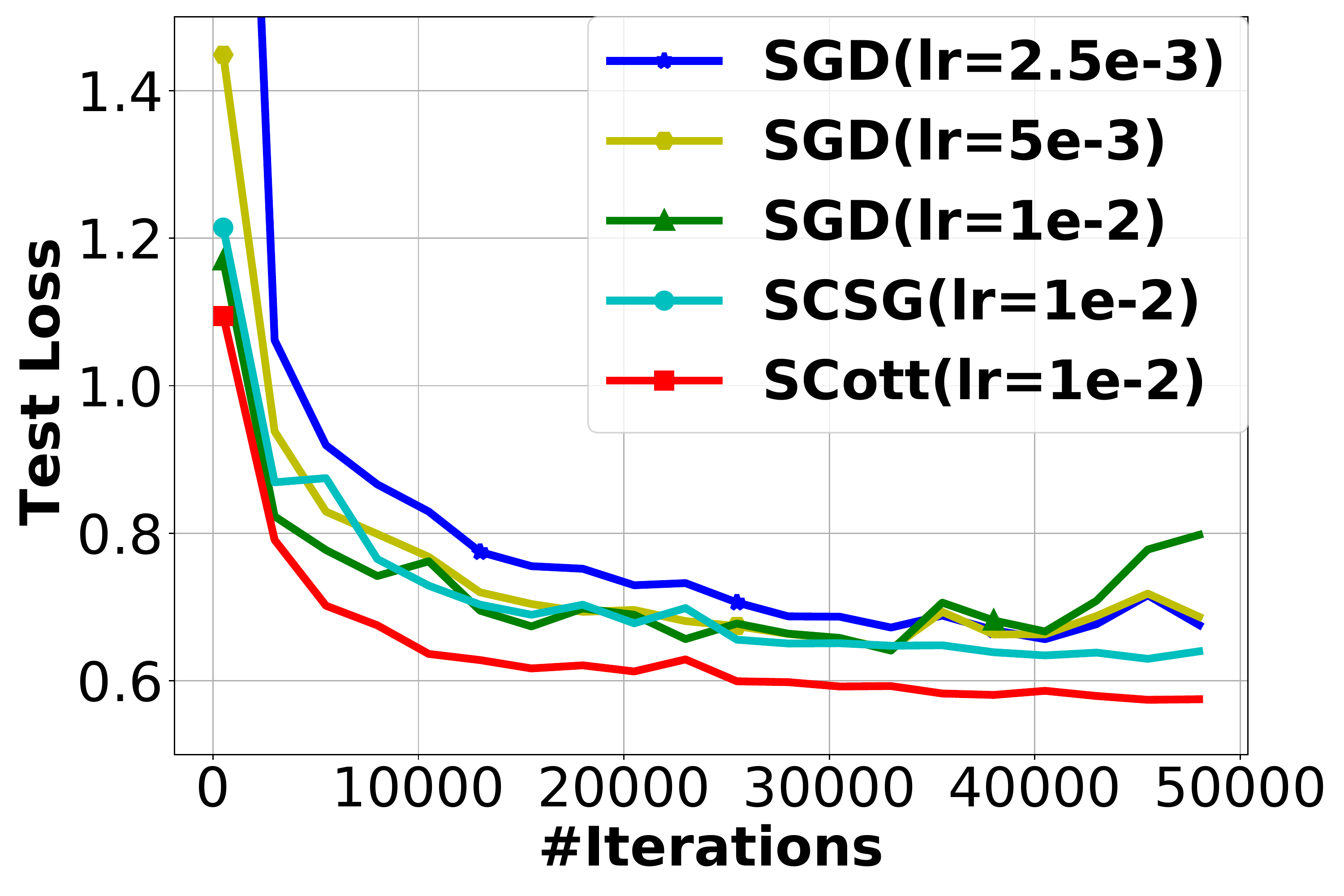}
        \label{nbeats_electricity_val}
    }
     \subfigure[Variance of Samplers \newline N-BEATS - Electricity Dataset]{
        \includegraphics[width=0.23\textwidth]{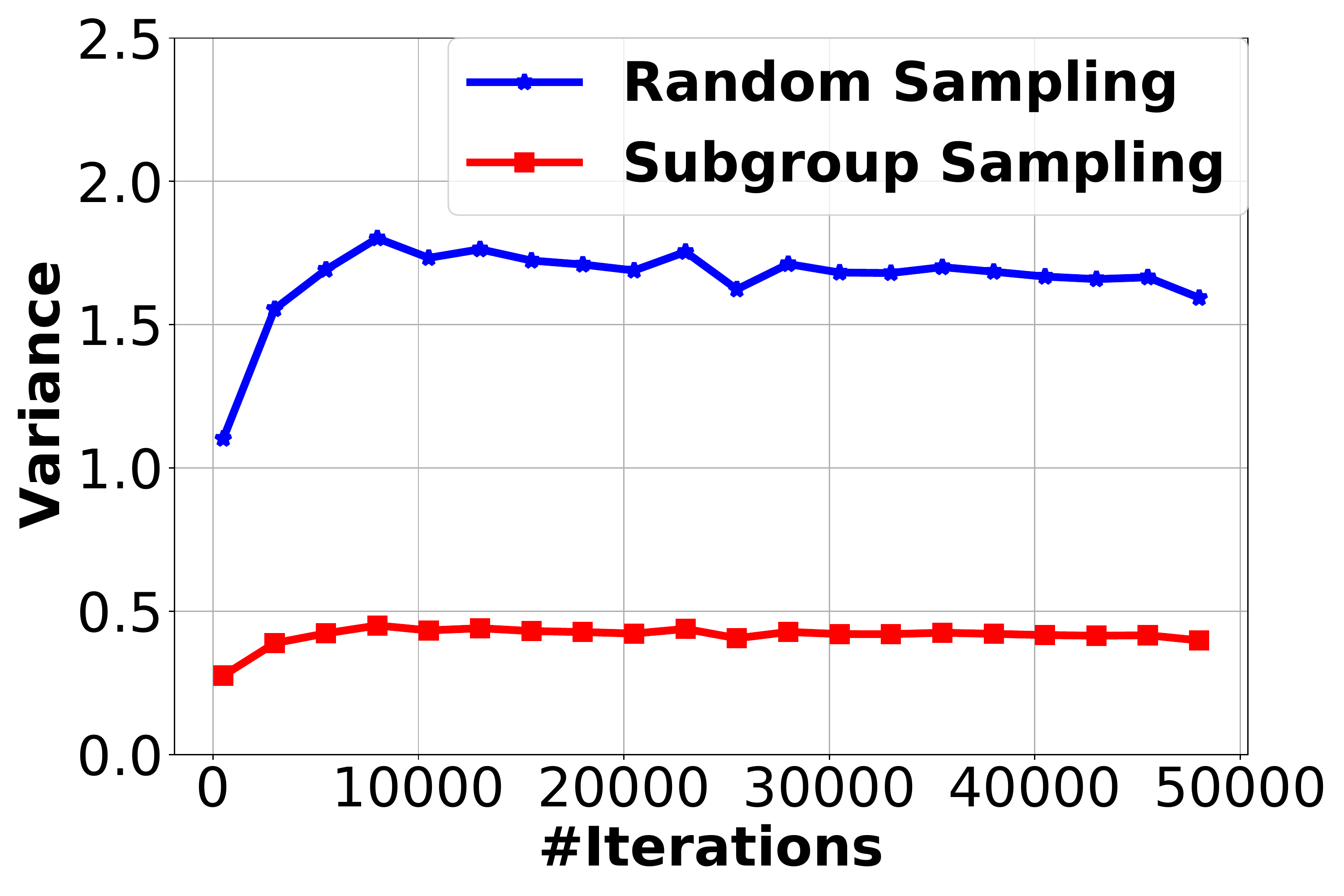}
        \label{nbeats_electricity_var}
    }
    %\vspace{-10pt}
    \caption{Performance of different optimizers. This experiment focus on comparing SCott with original SGD-type algorithms.}
\end{figure*}
%%%%%%%%%%Extension%%%%%%%%%%%%%%%%%%
\begin{figure*}[ht!]
    \subfigure[Training Loss / Time(s) \newline MLP - Traffic Dataset]{
        \includegraphics[width=0.23\textwidth]{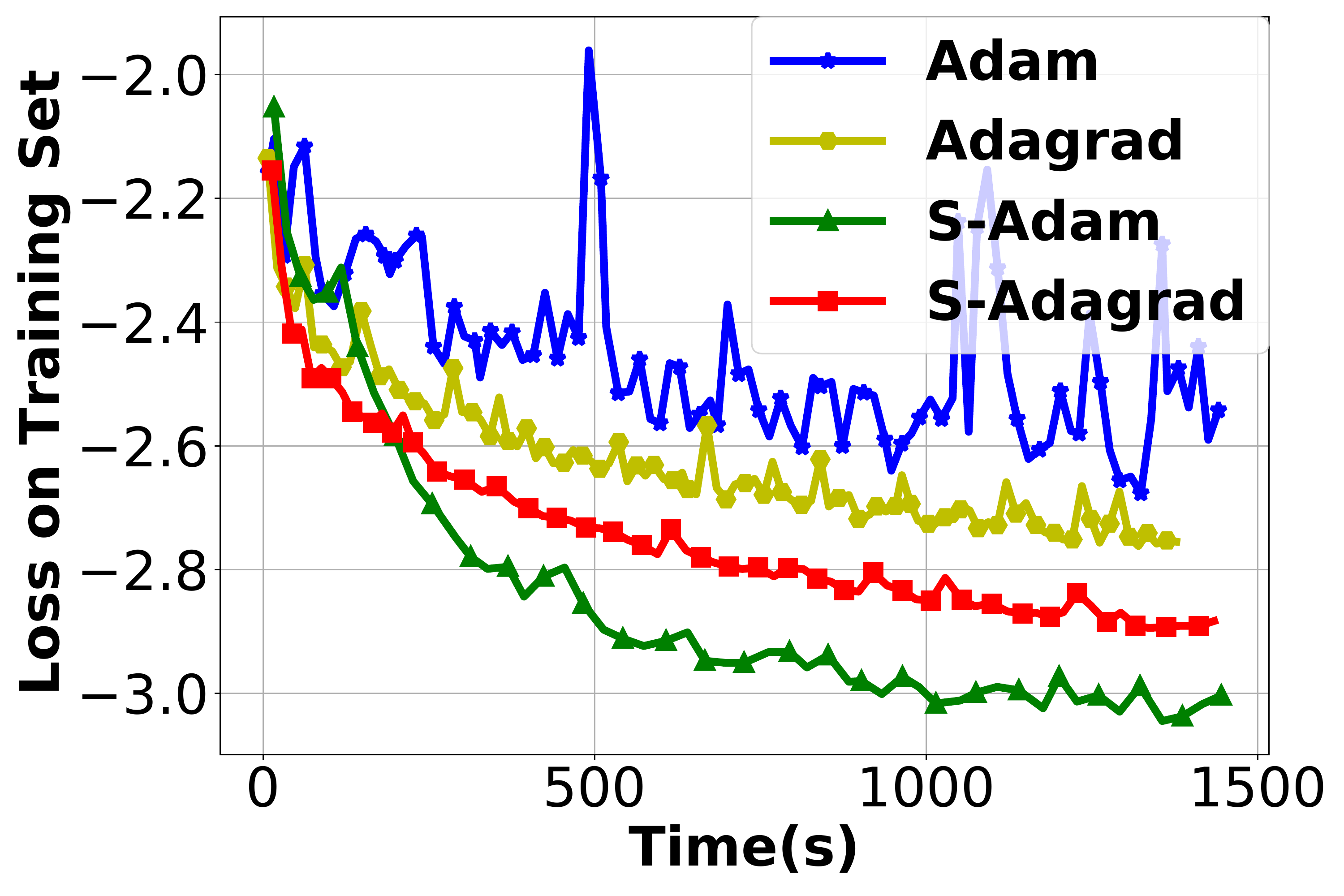}
    }
    \subfigure[Test Loss / Time(s) \newline MLP - Traffic Dataset]{
        \includegraphics[width=0.23\textwidth]{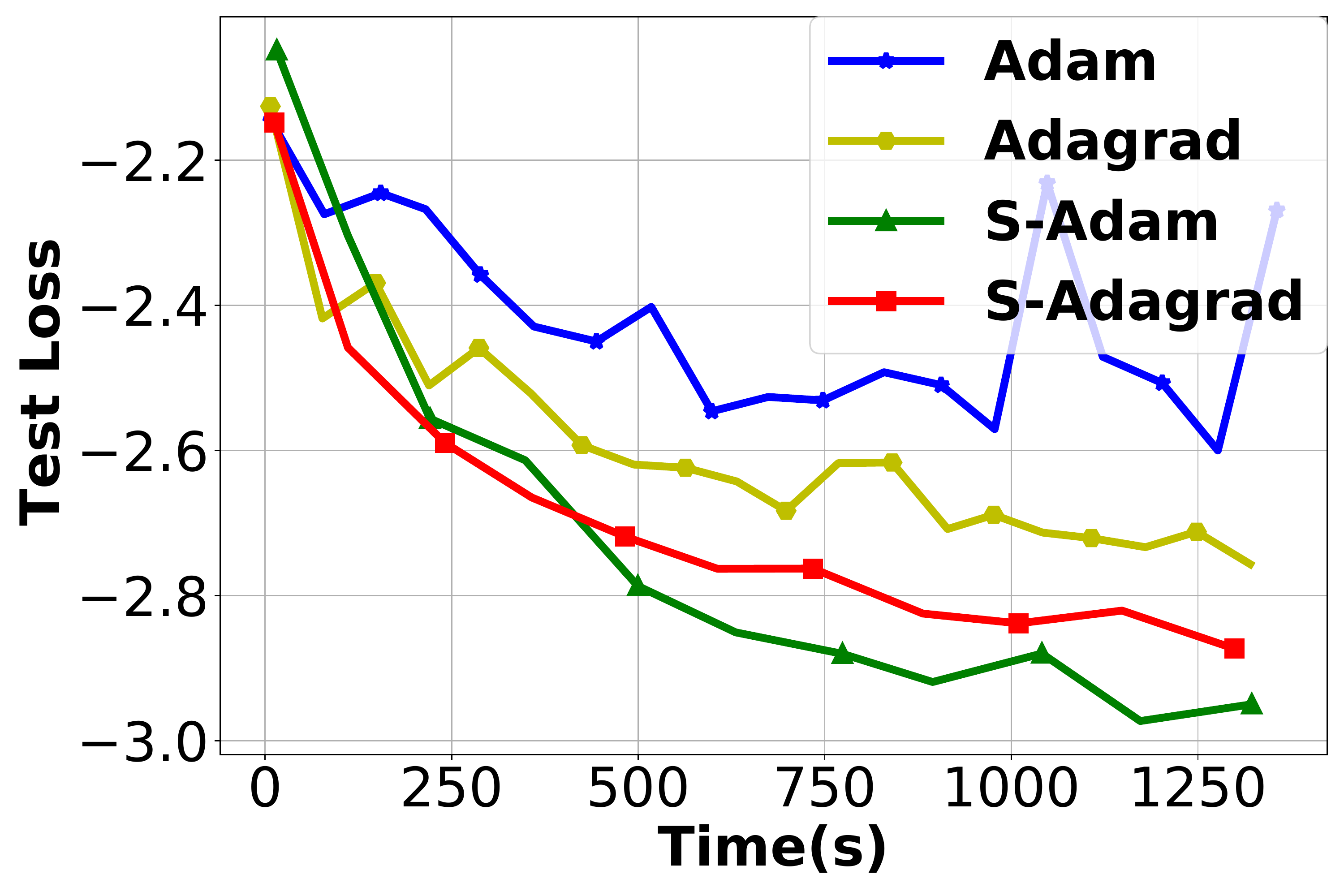}
    }
    \subfigure[Training Loss / Time(min) \newline N-BEATS - Electricity Dataset]{
        \includegraphics[width=0.23\textwidth]{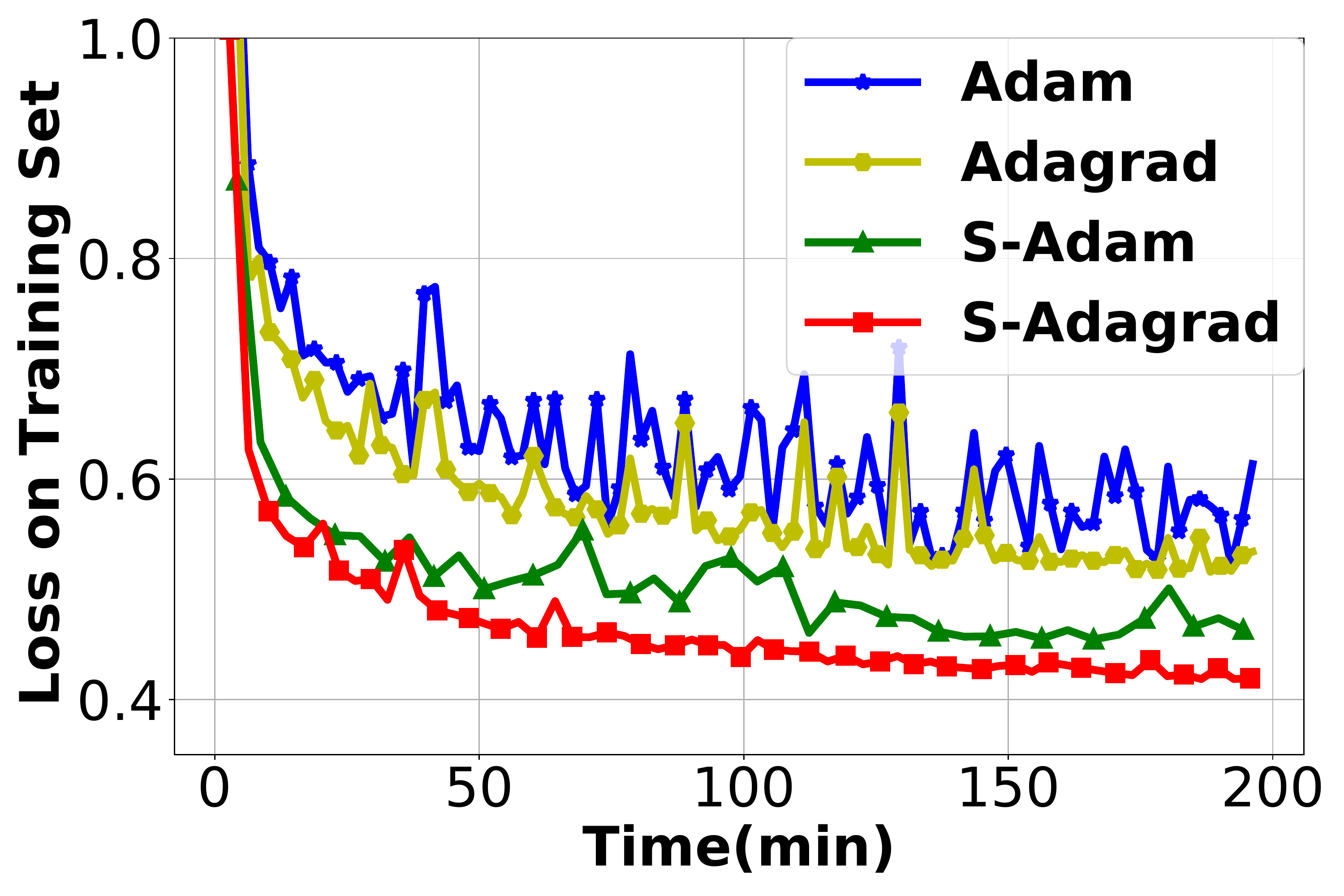}
    }
    \subfigure[Test Loss / Time(min) \newline N-BEATS - Electricity Dataset]{
        \includegraphics[width=0.23\textwidth]{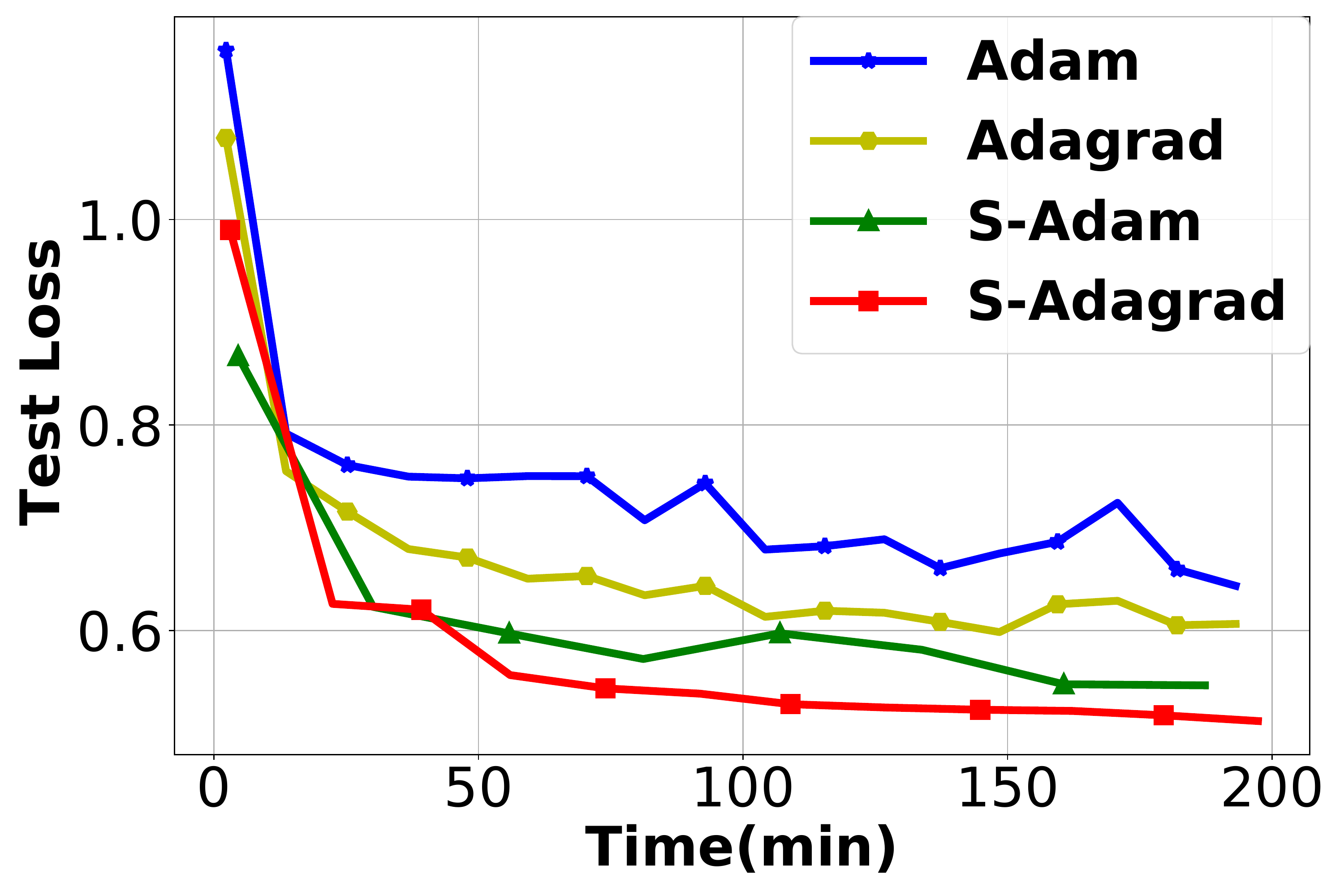}
    }
    \caption{SCott Variants on Adam and Adagrad. The "S-" denotes SCott version of the optimizer. The results are fine tuned.}
    \label{exp extension}
\end{figure*}

\section{Experiment}\label{experiment section}
In this section we empirically evaluate our algorithms and investigate how SCott improves optimizers in practice. We implement SCott in PytorchTS, a time series forecasting library \cite{rasul2020multi}. All the tasks run on a local machine configured with a 2.6GHz Inter (R) Xeon(R) CPU, 8GB memory and a NVIDIA GTX 1080 GPU.

In Section~\ref{section_synthetic_exp} and Section~\ref{section_real_exp}, we focus on comparing SCott with SGD and SCSG in different settings. These optimizers all have SGD-style update formulas, which helps us better understand the effect of variance reduction.
Additionally, in Section~\ref{section_extension_exp}, we modify the main step of SCott and let it follow the update rule of Adam and Adagrad. We rerun the experiments from previous sections on the two SCott variants.

\textbf{Models and Loss functions.}
In this section, we focus on four representative settings in forecasting problems:
\begin{itemize}[nosep,leftmargin=12pt]
    \item Vector AutoRegressive Model (VAR) with MSE loss \cite{lai2018modeling}.
    \item LSTM with MSE loss \cite{maharaj2019time}. We set the hidden layer size to be 100 and the depth to be 2.
    \item Simple FeedForward Network (MLP) with Negative Log Likelihood (NLL) loss \cite{alexandrov2019gluonts}. We set the hidden layer size to be 80 and the depth to be 4. 
    \item N-BEATS with MAPE loss \cite{oreshkin2019n}. We set the number of stacks to be 30.
\end{itemize}

\subsection{Warm-up: Synthetic Dataset}\label{section_synthetic_exp}
We start from training VAR and LSTM on a synthetic dataset, where we know the ground-truth for stratification. We generate the synthetic dataset by repeatedly transforming a linear curve based on simple functions such as \textit{sin} and \textit{polynomial}. For brevity, the details for generating the dataset is discussed in the supplementary material.
We plot the results of SGD and SCott from Figure~\ref{VAR_synthetic_iteration} to \ref{LSTM_synthetic_time}. We show SGD and SCott with the fine-tuned learning rate while for SGD we show the top three curves with different learning rate.
In Figure~\ref{VAR_synthetic_iteration} and Figure~\ref{LSTM_synthetic_iteration}. 
For SGD, if the learning rate is large, then the convergence curve is noisy and unstable. To ensure stable convergence, SGD needs to adopt small learning rate, and that results in more iterations. 
SCSG, on the other hand, slightly improve over SGD, while is still noisy in later iterations.
By comparison, SCott contains less variance noise, and it allows us to use larger learning rate while keeping a stable convergence.

\subsection{Real World Applications}\label{section_real_exp}
We proceed to discuss the performance on real-world applications. In this experiment, we train the FeedForward Network (MLP) and N-BEATS.
We use three public benchmark datasets: \textit{Traffic}, \textit{Exchange-Rate} and \textit{Electricity} \cite{lai2018modeling}, where details are shown as below:
\begin{itemize}[nosep,leftmargin=12pt]
    \item \textit{Traffic}: A collection of hourly data from the California Department of Transportation. The data describes the road occupancy rates (between 0 and 1) measured by different sensors on San Francisco Bay area free ways. For this dataset, we set $\tau_c=3$ days (72 hours) and $\tau_p=1$ day (24 hours).
    \item \textit{Exchange-Rate}: the collection of the daily exchange rates of eight foreign countries including Australia, British, Canada, Switzerland, China, Japan, New Zealand and Singapore ranging from 1990 to 2016. For this dataset, we set $\tau_c=8$ days and $\tau_p=1$ day.
    \item \textit{Electricity}: The electricity consumption in kWh was recorded hourly from 2012 to 2014, for n = 321 clients. We converted the data to reflect hourly consumption. For this dataset, we set $\tau_c=3$ days (72 hours) and $\tau_p=1$ day (24 hours).
\end{itemize}

\textbf{Stratification.}
As mentioned in Section~\ref{sec:stratified_training},
we adopt a simple timestamp-based stratification policy for all the datasets.
For \textit{Traffic} dataset: we stratify all the time series segments only based on its weekday and season, i.e., two time series segments are in the same stratum if and only if their weekday and season are the same. This results in total 49 subgroups, we repeat this process on \textit{Electricity} dataset and it results in 70 strata; For \textit{Exchange-Rate} dataset, we first evenly divide the whole time series into 6 range based on the time stamps. And then within each range we further group the time series segments based on the time series instance, this results in total 32 strata.

\textbf{Results Analysis.}
We first see from Table~\ref{table experiment} that given same time budget, SCott achieves smaller loss on both training and testset. We further
plot the training curves in Figure~\ref{MLP_traffic_iteration} to \ref{MLP_traffic_time} and Figure~\ref{nbeats_electricity_iteration} to \ref{nbeats_electricity_time}. We can observe the results are mostly aligned with our results on synthetic dataset: with stratified sampling, SCott is able to adopt large learning rate which allows it to converge faster compared to SGD and SCSG.
Additionally, we verify in Figure~\ref{MLP_traffic_val} and \ref{nbeats_electricity_val} the benefits of SCott does not compromise the validation error of the model. Finally, we plot in Figure~\ref{MLP_traffic_var} and \ref{nbeats_electricity_var} that the stratified sampling does induce smaller variance compared to uniform sampling, even with the simple stratification policy we adopt.

%\vspace{-8pt}
\subsection{Variants of SCott}\label{section_extension_exp}
%%\vspace{-10pt}
So far, we focus on comparing SCott with SGD and SCSG. Moreover, we investigate how SCott can be applicable to enhance other types of optimizers, Adam and Adagrad. To do so, we incorporate the main step of SCott in Equation~(\ref{SCott main step}) into the update rule of Adam and Adagrad, which we refer to as S-Adam and S-Adagrad.
We rerun all the experiments on the real-world dataset.
We first see from Table~\ref{table experiment} S-Adam/S-Adagrad is able to achieve smaller loss given same time.
Then we further plot the training curves in Figure~\ref{exp extension}.
We find SCott is able to improve both Adam and Adagrad by a certain margin. As shown in Table~\ref{table experiment}, these SCott variants improve upon their non-SCott baselines without compromising the test loss.
Additionally, comparing SGD-type with Adam- and Adagrad-type optimizers, SCott sometimes can outperform Adam and Adagrad (such as in MLP on traffic dataset). On the other hand, we find S-Adam and S-Adagrad consistently outperform SGD-type optimizers as well as Adam and Adagrad.

%\vspace{-10pt}
\section{Conclusions}
%%\vspace{-10pt}
In this paper, we show that heterogeneity in large scale time series data is detrimental to the convergence of the stochastic optimizers. To address the challenge, we introduce SCott, a variance reduced optimizer that speeds up the training of forecasting models based on stratified time series data. A novel convergence analysis is provided for SCott, which by varying the stratification conditions, recovers the well-known results in stochastic optimization. Empirically, we show SCott converges faster compared to plain stochastic optimizer, with respect to both iterations and time on both synthetic and real-world dataset. We leave the future works of investigating the effect of stratification on SCott variants, applying SCott tasks beyond forecasting, and developing practical stepsize selection \cite{park2020variable,yu2020variable}.

\section*{Acknowledgement}
The authors would like to thank Valentin Flunkert, Konstantinos Benidis, Tim Januschowski, Syama Sundar Rangapuram, Lorenzo Stella, Laurent Callot and anonymous reviewers from ICML 2021 for providing valuable feedbacks on the earlier versions of this paper.

\bibliography{main}
\bibliographystyle{icml2021}

\newpage
\appendix
\onecolumn
\icmltitle{Variance Reduced Training with Stratified Sampling \\
for Forecasting Models \\ (Supplementary Materials)}
\section{Details in Experiments and Addtional Results}\label{section appendix exp}
\subsection{Practical Version of SCott}\label{sec:app:earlystop}
We first discuss in details on the pratical version of SCott as mentioned in Section~\ref{application section}. 
The full description is shown in Algorithm~\ref{practical svrggroup}. The main difference between the plain SCott and Algorithm~\ref{practical svrggroup} is the latter treats the period $K$ of performing stratified as a constant, while adaptively altering $K$ based on a stopping criteria $\gamma$. 
Adopting such technique, despite being complicated in theory, allows us to adaptively perform the stratified sampling based on the progress of the training. The hyperparameter $\gamma$ can then be obtained via standard tuning algorithms such as grid search and random search.

\begin{algorithm}[ht]
%\begingroup
%    \setlength{\abovedisplayskip}{1pt}
%    \setlength{\belowdisplayskip}{1pt}
	\caption{The practical version of SCott, where we apply the early stopping technique, instead of choosing $K$ based on Geometric distribution.}\label{practical svrggroup}
	\begin{algorithmic}[1]
		\REQUIRE Total number of iterations $T$, learning rate $\{\alpha_t\}_{1\leq t\leq T}$, initialize $\*\theta^{(0,0)}$, strata: $\{\mathcal{D}_i\}_{1\leq i\leq B}$, initialized selection of $K$, hyperparameter $\gamma$.
		
		%for
		\FOR{$t=0,1,\cdots, T-1$}
			\STATE  Sample a $\xi_i^{(t)}$ from stratum $i$ and perform stratified sampling (with $w_i=|\mathcal{D}_i|/|\mathcal{D}|$): \newline $\*{g}^{(t,0)}=\sum\nolimits_{i=1}^{B}w_i\nabla f_{\xi_i^{(t)}}(\*\theta^{(t,0)})$.
			\FOR{$k=0,1,\cdots,K-1$}
			    \STATE Sample $\xi^{(t,k)}$ from $\mathcal{D}$.
			    \STATE Compute the update $\*v^{(t,k)}$ as $\nabla f_{\xi^{(t,k)}}(\*\theta^{(t,k)})-\nabla f_{\xi^{(t,k)}}(\*\theta^{(t,0)})+\*g^{(t,0)}$.
			    \STATE Update the parameters as $\*\theta^{(t,k+1)} = \*\theta^{(t,k)} - \alpha_t\*v^{(t,k)}$.
			    \IF{$\|\*v^{(t,k)}\|^2 \leq \gamma \|\*v^{(t,0)}\|^2$}
			        \STATE \textbf{break}
			    \ENDIF
		    \ENDFOR
		    \STATE Set $\*\theta^{(t+1,0)}=\*\theta^{(t,k+1)}$.
		\ENDFOR
		\STATE \textbf{return} Sample $\hat{\*\theta}^{(T)}$ from $\{\*\theta^{(t,0)}\}_{t=0}^{T-1}$ with $\mathbb{P}(\hat{\*\theta}^{(T)}=\*\theta^{(t,0)})\propto\alpha_tB$
	\end{algorithmic}
%	\endgroup
\end{algorithm}

\subsection{Synthetic Dataset Generation}
In this subsection, we introduce the details of generating synthetic time series dataset as used in the experiments.
We first set the context length to be 72 and prediction length to be 24.
This synthetic dataset contains 4 time series with different patterns on their time horizon. We start from the definition of four types of patterns:
$\*t = [1, 2, \cdots, 23, 24]\in\mathbb{R}^{24},\hspace{0.5em} P_1 = \sin(\*t)\in\mathbb{R}^{24}, \hspace{0.5em}  P_2 = \*t\in\mathbb{R}^{24}, \hspace{0.5em}P_3 = \*t^2\in\mathbb{R}^{24}, \hspace{0.5em}  P_4 = \sqrt{\*t}\in\mathbb{R}^{24}$,
where all the transformations are element-wise, e.g. $\sin(\*t) = [\sin(1), \sin(2), \cdots, \sin(23), \sin(24)]$. And the four patterns are four different time series slices of length 24 that maps $\*t$ to different values via transformations of \textit{sin}, \textit{linear}, \textit{quadratic}, \textit{square root}, respectively. With these patterns, the time series data are then constructed via concatenating the patterns with different orders on the time horizon.
Specifically,
\begin{align*}
    & TS_1 = \underbrace{[P_1, P_2, P_3, P_4, \cdots, P_1, P_2, P_3, P_4]}_{[P_1, P_2, P_3, P_4]\text{ repeats 2K times}} + \mathcal{N}(0, \*1), \hspace{1em}\text{ where }\hspace{1em} \mathcal{N}(0, \*1)\in\mathbb{R}^{192K}\\
    & TS_2 = \underbrace{[P_4, P_3, P_2, P_1, \cdots, P_4, P_3, P_2, P_1]}_{[P_4, P_3, P_2, P_1]\text{ repeats 2K times}} + \mathcal{N}(0, \*1), \hspace{1em}\text{ where }\hspace{1em}\mathcal{N}(0, \*1)\in\mathbb{R}^{192K}\\
    & TS_3 = \underbrace{[P_1, P_3, P_2, P_4, \cdots, P_1, P_3, P_2, P_4]}_{[P_1, P_3, P_2, P_4]\text{ repeats 2K times}} + \mathcal{N}(0, \*1), \hspace{1em}\text{ where }\hspace{1em}\mathcal{N}(0, \*1)\in\mathbb{R}^{192K}\\
    & TS_4 = \underbrace{[P_4, P_2, P_3, P_1, \cdots, P_4, P_2, P_3, P_1]}_{[P_4, P_2, P_3, P_1]\text{ repeats 2K times}} + \mathcal{N}(0, \*1), \hspace{1em}\text{ where }\hspace{1em}\mathcal{N}(0, \*1)\in\mathbb{R}^{192K}
\end{align*}
where $\mathcal{N}(0, \*1)$ denotes a random vector where each coordinate is sampled from a normal distribution.
We can see each time series is repeating a distinct order of the four patterns, forming a temporal pattern, on its time horizon. We let such temporal pattern repeat 2K times on the time horizon of each time series. Finally, a Gaussian noise is added to each time series to capture randomness. 
%A snippet of the generated time series is shown in Figure~\ref{linear_dataset}.
After generating the time series, we then extract the training examples via a sliding window which slides 24 timestamps\footnote{We do this to guarantee each training example can include a completely different pattern in its prediction range. In practice, this can mean extracting training examples by day on a hourly measured time series.} between adjacent training examples. With a simple calculation, the total number of training examples in this dataset is 32K.

\paragraph{Stratification.}
From the data generation process, we can see in each time series contains exactly four types of mapping: take $TS_1$, the type of mappings on its time horizon are $\forall i=1,2,3,4$,
\begin{align*}
    &\text{Mapping i: }[P_{(i+1)\bmod 4}, P_{(i+2)\bmod 4}, P_{(i+3)\bmod 4}]\rightarrow P_i
\end{align*}
Same conclusions can be drawn on other time series. And then we can stratify all the training examples via a  simple judgemental policy: the training examples that belonging to the same time series and having same pattern in their prediction range are clustered to the same stratum. The total number of strata is then 16.

\subsection{Additional Results}

\subsubsection{Results additional to the main paper settings}
In the main paper, we show the convergence curves of training MLP on Traffic dataset and training NBEATS on Electricity dataset. Here, we provide the additional curves of training MLP on Electricity dataset.
%%%%%%%%%%Exchange Rate Dataset%%%%%%%%%%%%%%%%%%
\begin{figure*}[ht]
%\centering
    \subfigure[Training Loss / \# Iterations \newline MLP - Electricity Dataset]{
        \includegraphics[width=0.23\textwidth]{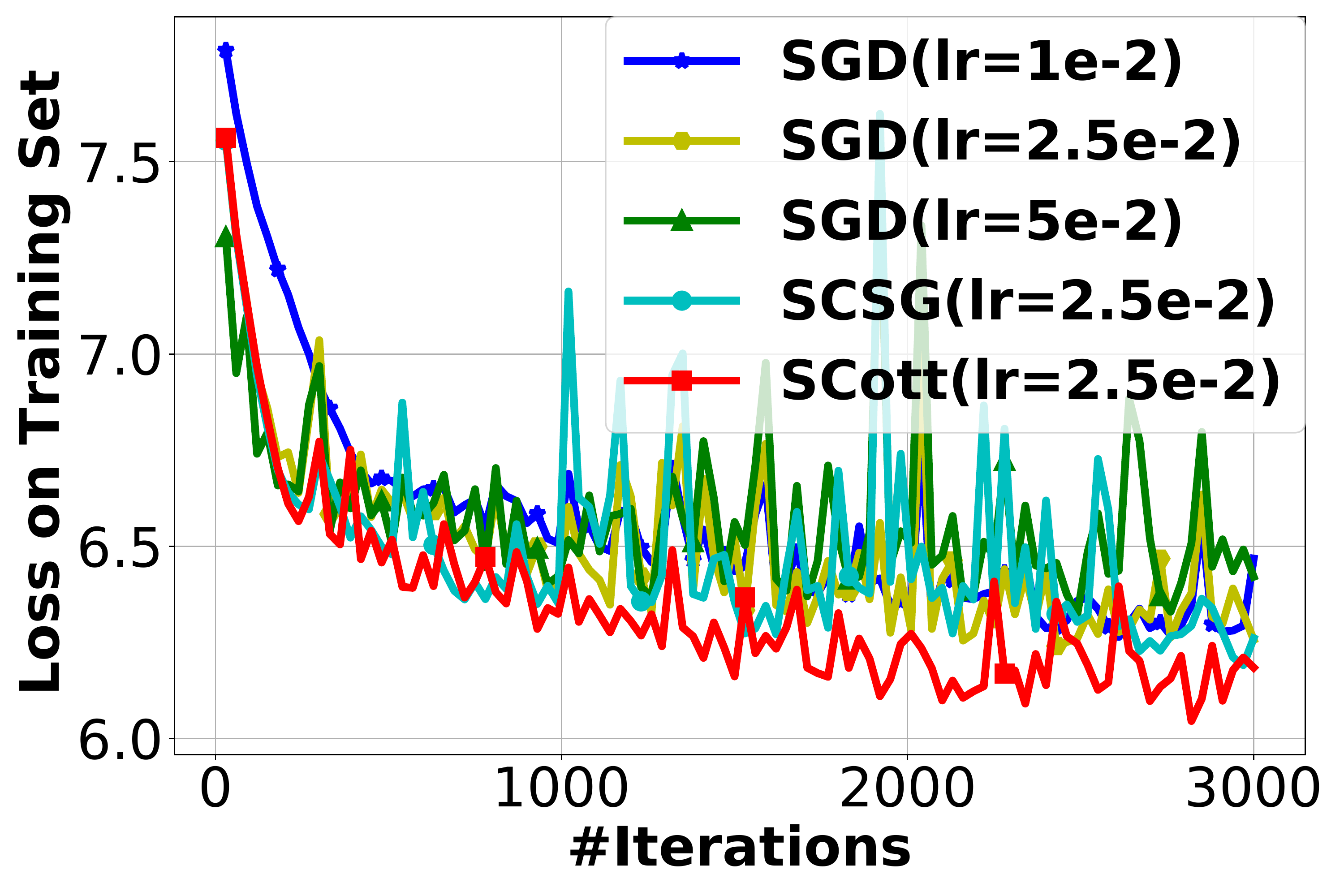}
        \label{mlp_electricity_iteration}
    }
     \subfigure[Training Loss / Time(s) \newline MLP - Electricity Dataset]{
        \includegraphics[width=0.23\textwidth]{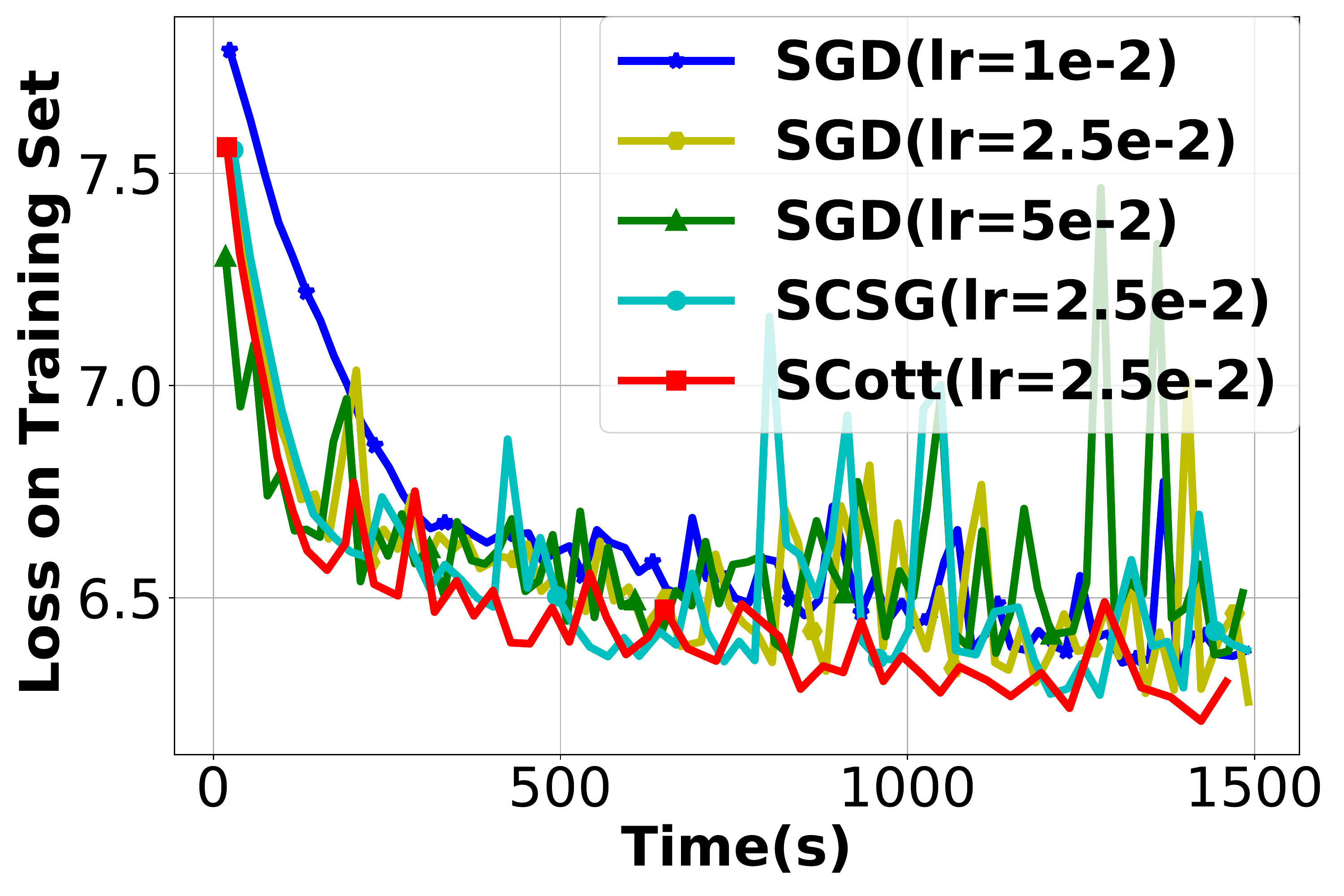}
        \label{mlp_electricity_time}
    }
%\centering
    \subfigure[Test Loss / \# Iterations \newline MLP - Electricity Dataset]{
        \includegraphics[width=0.23\textwidth]{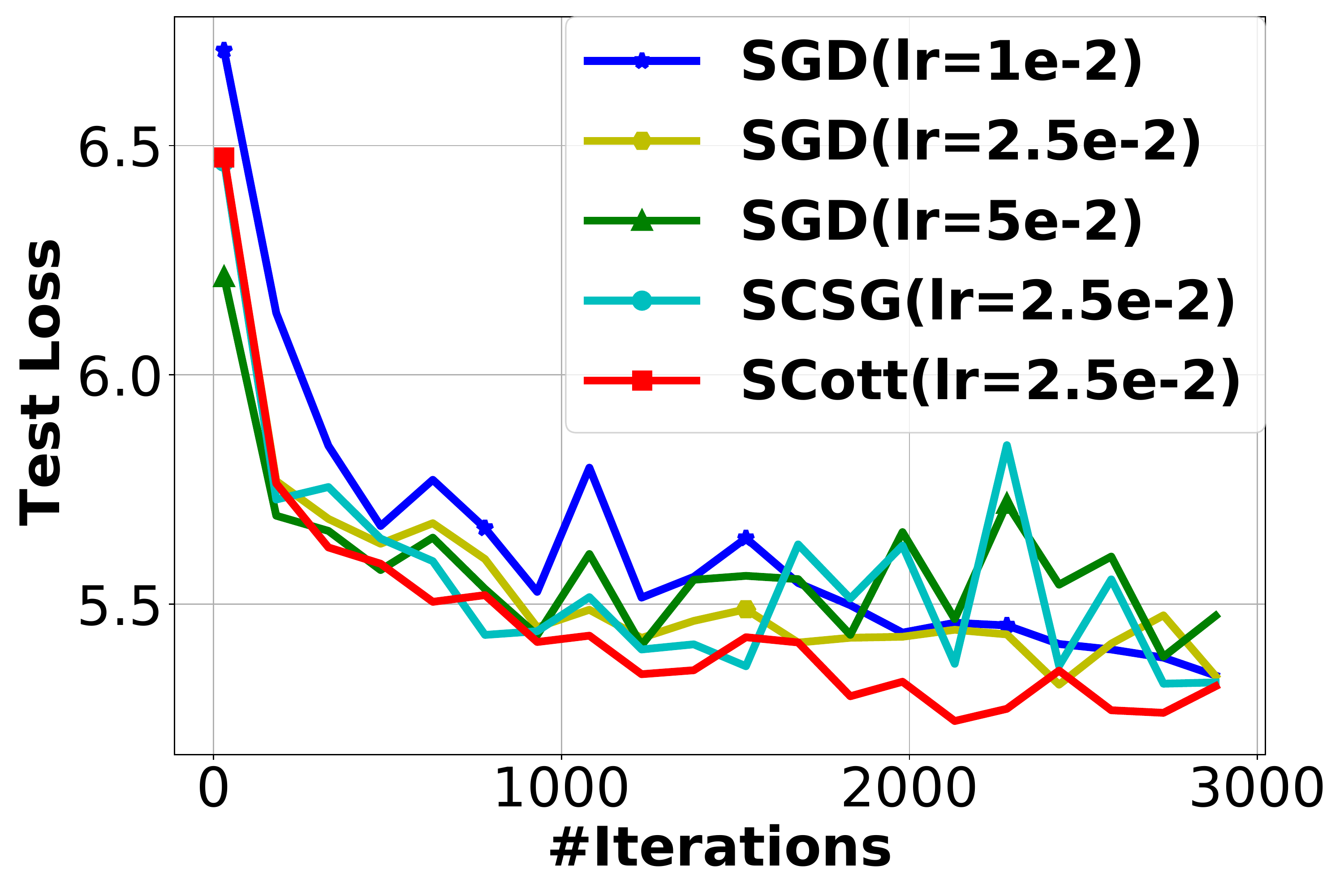}
        \label{mlp_electricity_val}
    }
     \subfigure[Variance of Samplers \newline MLP - Electricity Dataset]{
        \includegraphics[width=0.23\textwidth]{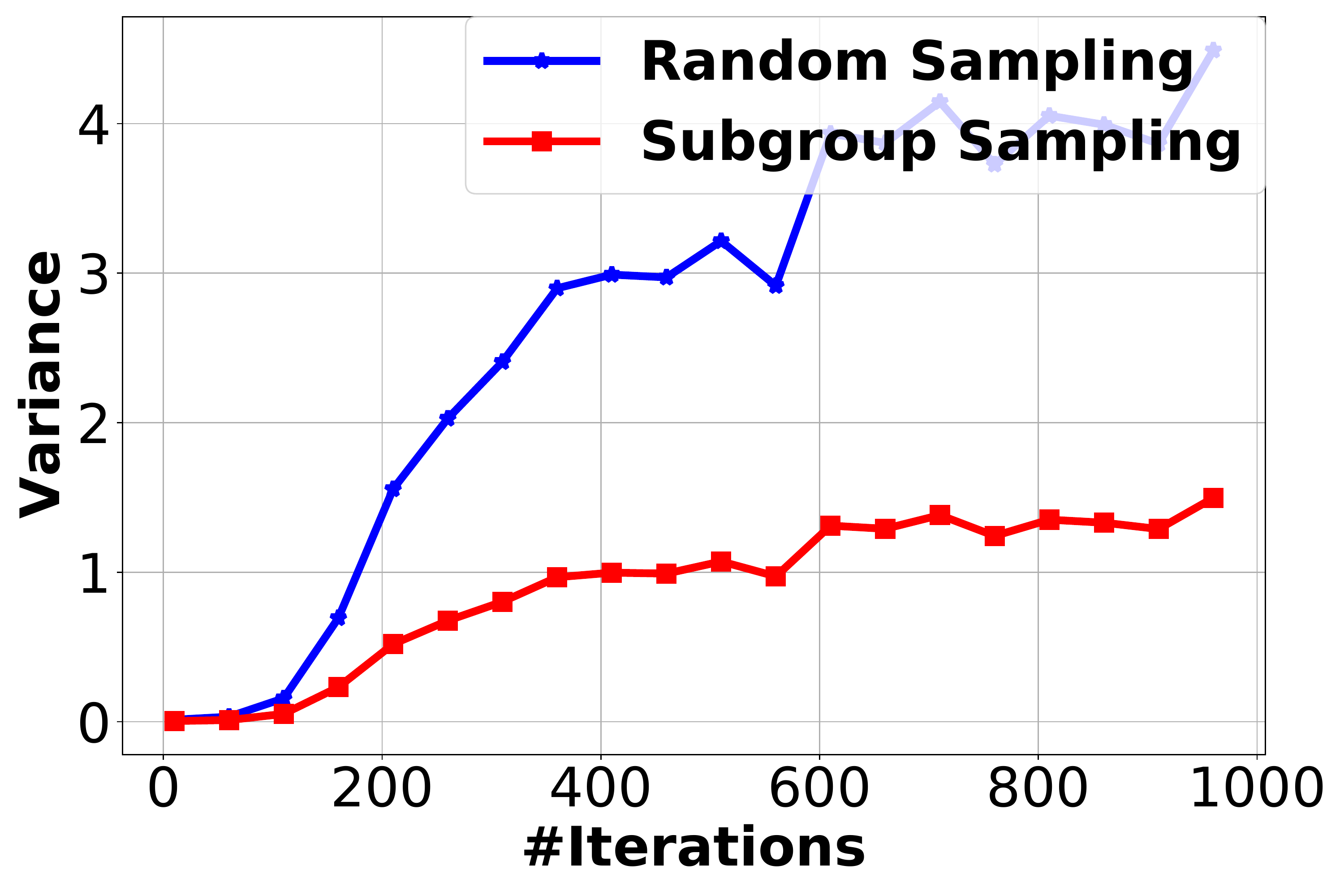}
        \label{mlp_electricity_var}
    }
    \caption{Additional Results of MLP on Electricity Dataset.}
\end{figure*}

\subsubsection{Applying Early Stopping Technique to SCSG and SVRG}
\begin{figure*}[ht]
\centering
    \subfigure[Training Loss w.r.t. Time(s) \newline Traffic Dataset]{
        \includegraphics[width=0.23\textwidth]{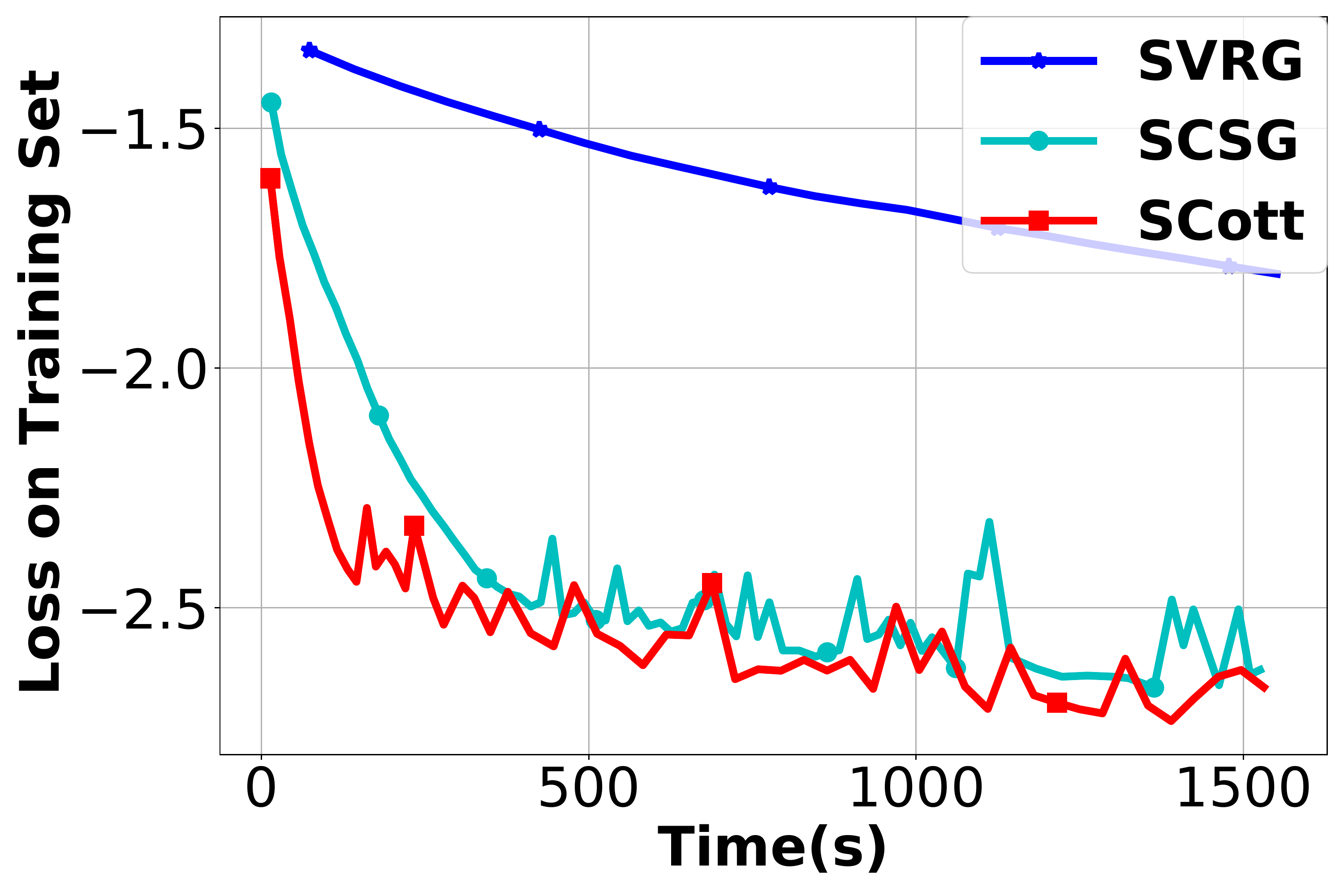}
    }
    \subfigure[Test Loss w.r.t. Time(s) \newline Traffic Dataset]{
        \includegraphics[width=0.23\textwidth]{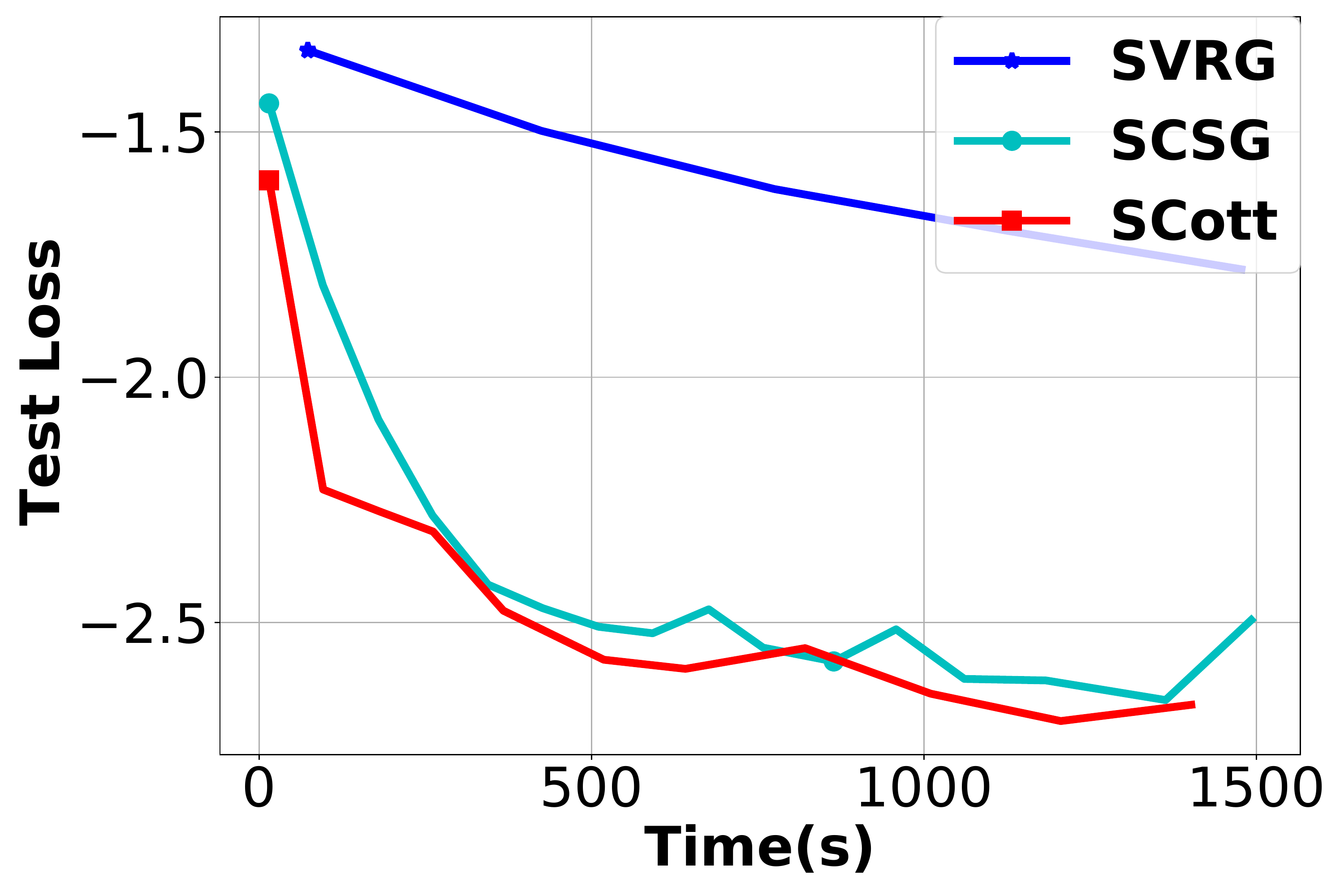}
    }
    \subfigure[Training Loss w.r.t. Time(s) \newline Electricity Dataset]{
        \includegraphics[width=0.23\textwidth]{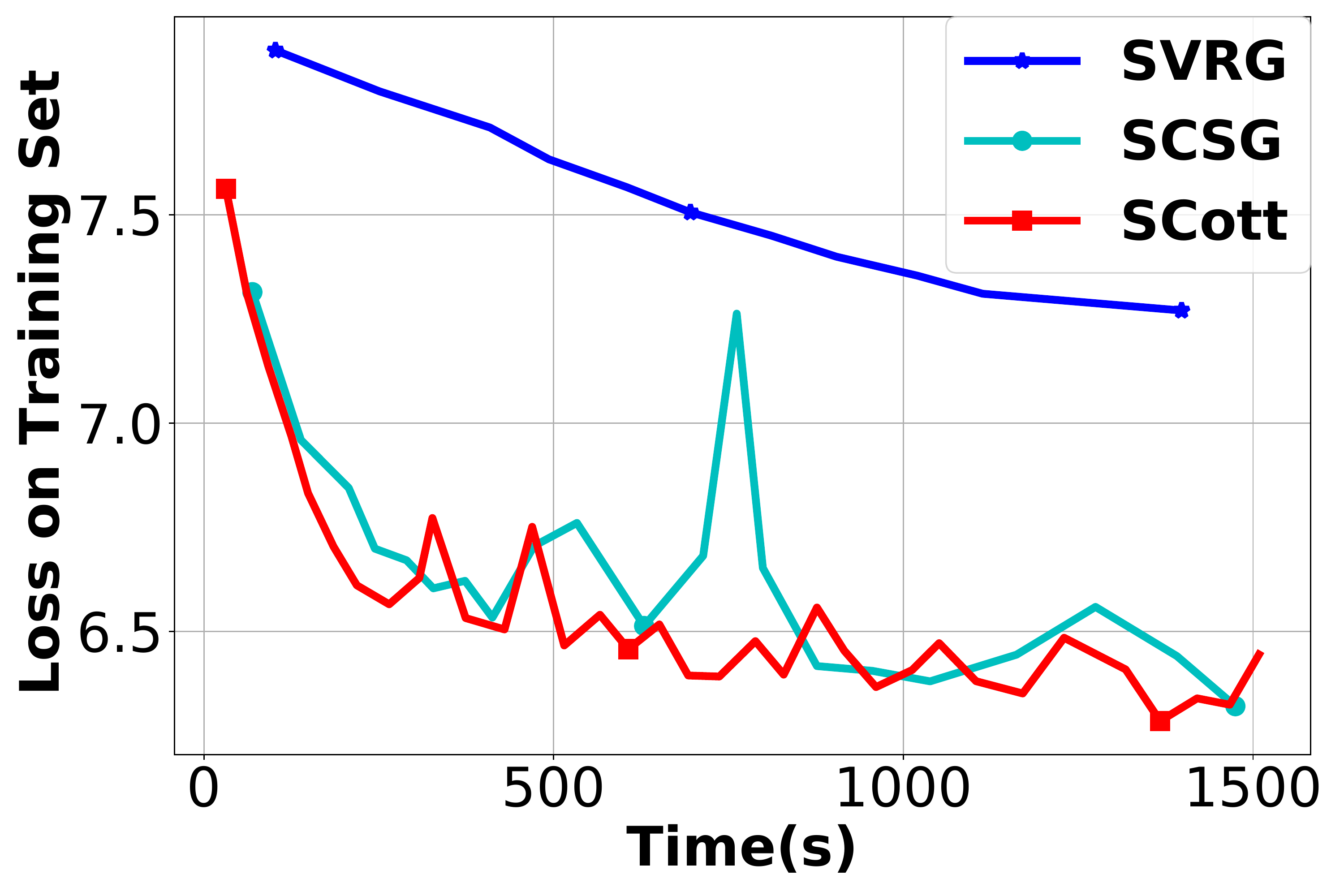}
    }
    \subfigure[Test Loss w.r.t. Time(s) \newline Electricity Dataset]{
        \includegraphics[width=0.23\textwidth]{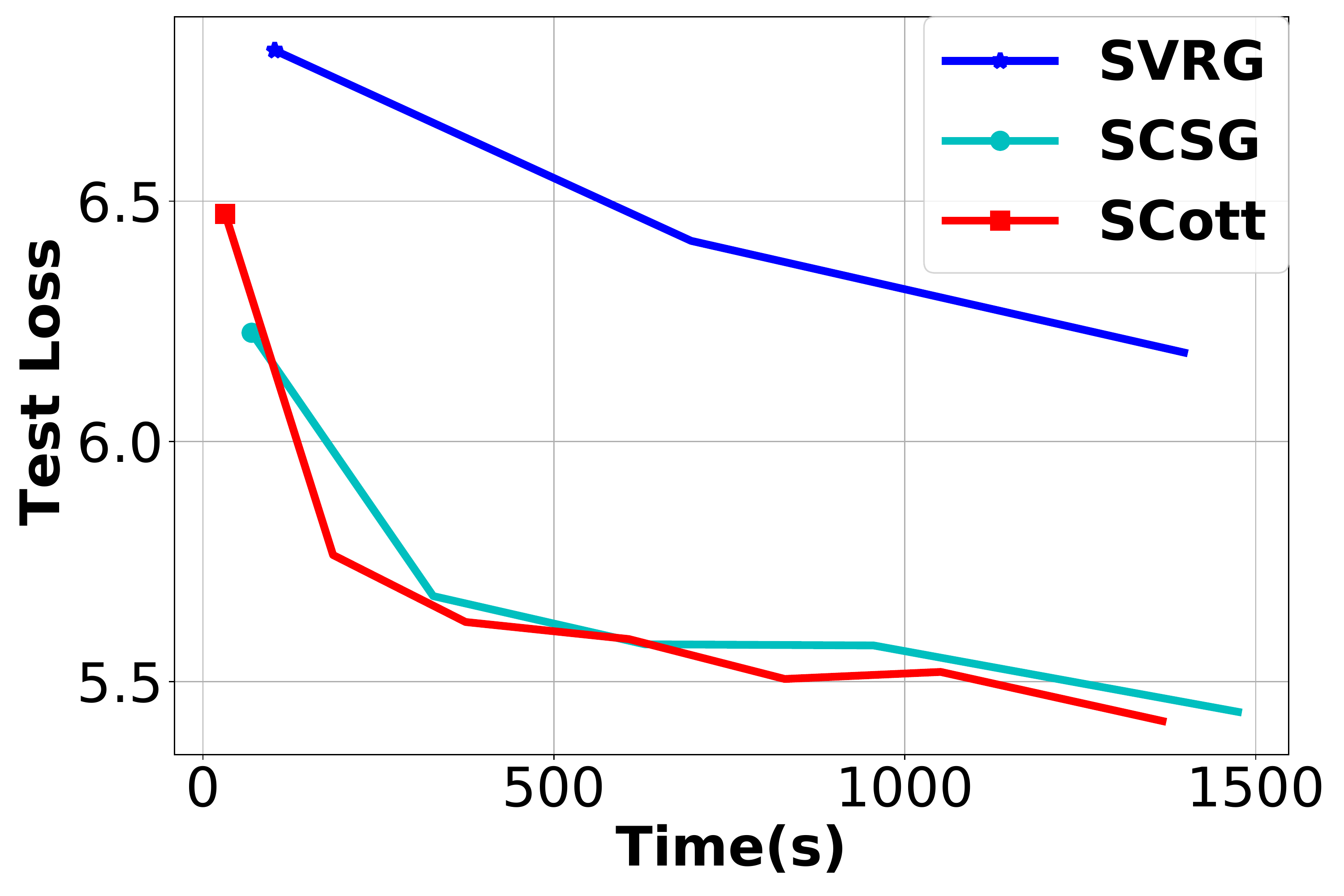}
    }
    \caption{Applying the early stopping technique on SCSG and SVRG.}
    \label{fig:app:gamma_trick}
\end{figure*}

In this subsection, we investigate how the baseline SCSG and SVRG would perform when the early stopping technique introduced in Section~\ref{sec:app:earlystop} is applied on them. We rerun the MLP model on Traffic and Electricity dataset, and the fine tuned results are shown in Figure~\ref{fig:app:gamma_trick}. In the literature, SVRG is shown to perform bad on deep learning tasks \cite{defazio2018ineffectiveness}.

\subsection{Hyperparameter Tuning}
We apply the grid search to tune the hyperparameters in each experiment, the grids for the step sizes are:
\{0.1, 0.05, 0.025, 0.01, 0.005, 0.0025, 0.001, 0.0005, 0.00025, 0.0001\}. We set the weight decay to be 1e-5.
For experiments in Section~\ref{section_synthetic_exp} and \ref{section_real_exp}, the optimal step size is specified. For SCott, we additionally tune $\gamma$ from $\{0.1, 0.125, 0.15, 0.2\}$, and the optimal choice of $\gamma$ on synthetic dataset is 0.125. Hyperparameters in other settings is shown in Table~\ref{table hyperparameter}.
For the experiment in section~\ref{section_extension_exp}, we have two extra hyperparameters $\beta_1$ and $\beta_2$, we set the grids to be \{0.9, 0.99, 0.999\}. Finally, for $\gamma$, we extend the grids range to \{0.1, 0.125, 0.15, 0.2, 0.4, 0.8\}.
The optimal choice of hyperparameter for each settings are shown in Table~\ref{table hyperparameter}.

%\subsection{A Concrete example of pre-grouping with long- and short-term temporal pattern.}
%In the main paper, we discuss naively using the timestamp to pre-group the time series examples. Here we give a concrete pseudo code for such strategy, allowing pre-grouping with arbitrary long- and short-term temporal pattern \cite{lai2018modeling}. Denote the temporal patterns available as $\{\Delta_i\}_{i=1}^m$. We know that each training example extracted is placed at a certain time point within each temporal interval, and we can use such information as features and apply a random hashing grouping on top of it. The detailed description for such algorithm is shown in Algorithm~\ref{algo grouping}. Note that in the main paper, the pre-grouping strategy for \textit{Traffic} and \textit{Electricity} dataset can be seen as a special case of Algorithm~\ref{algo grouping} with two temporal intervals: weeks and seasons.

\begin{table}[ht]
\small
\label{table hyperparameter}
\caption{Hyperparameters used for experiments on real-world applications. The format of the hyperparameters is shown as the following: the first value is the optimal choice of step size. For SCott-type optimizers, the last value is the optimal choice for $\gamma$. Additionally, the $\beta_1$ and $\beta_2$ for Adam-type optimizers are set to be 0.9 and 0.999 respectively for optimal performance.}
\small
\begin{center}
\begin{tabular}{lll|l|l|l|l|l|l}
\toprule
\multirow{2}{*}{ Model } & \multirow{2}{*}{ Dataset } & \multicolumn{7}{c}{ Optimizer } \\
\cmidrule{3-9}
& & SGD & SCSG & SCott & Adam & S-Adam & Adagrad & S-Adagrad \\
\midrule
\multirow{3}{*}{ MLP } & Exchange Rate & 5e-3 & 5e-2 & 5e-2/0.125 & 5e-3 & 5e-3/0.1 & 2.5e-2 & 2.5e-2/0.1 \\
& Traffic & 2.5e-2 & 2.5e-2 & 2.5e-2/0.125 & 5e-3 & 5e-3/0.125 & 5e-3 & 5e-3/0.125 \\
& Electricity & 2.5e-2 & 2.5e-2 & 2.5e-2/0.125 & 5e-3 & 2.5e-3/0.125 & 5e-3 & 2.5e-3/0.125 \\
\midrule
\multirow{3}{*}{ NBEATS } & Exchange Rate & 1e-3 & 1e-3 & 1e-3/0.1 & 1e-3 & 1e-3/0.1 & 1e-3 & 1e-3/0.1 \\
& Traffic & 1e-4 & 1e-4 & 2.5e-3/0.125 & 1e-4 & 1e-4/0.125 & 1e-4 & 1e-4/0.125 \\
& Electricity & 5e-3 & 1e-2 & 1e-2/0.125 & 1e-2 & 1e-2/0.125 & 1e-2 & 1e-2/0.125 \\
\bottomrule
\end{tabular}
\end{center}
\end{table}
%\newpage
\section{Technical Proof.}
\subsection{Heterogeneity Noise with Uniform Sampling}\label{section heterogeneous error}
As a supplementary to the main paper, here we investigate another example to illustrate why time series data can be heterogeneous, and how uniform sampling could cause extra noise on such dataset.
Consider the simplest AR model with $p=1$. Without the loss of generality, we assume the parameter is initialized at point $0$: $\*\theta^{(0)}=0\in\mathbb{R}$.
Now consider a dataset $\mathcal{D}$ contains a single time series that takes the following form:
%%\vspace{-10pt}
\begin{align*}
    \underbrace{-1}_{t=1}, -\delta, 1, \delta, \underbrace{-1, -\delta, 1, \delta}_{\text{Temporal Pattern}}, \cdots,
\end{align*}
where $\delta>0$ denote some constant.
We can see in this example a temporal pattern of length $4$ is repeating on the time horizon, and the conditional distribution over the timestamp has two types.
With some simple analysis, for all the examples whose prediction time $t_0$ fulfilling $t_0\bmod 2=1$, their global minima are centering around $\theta_1^*=\delta$. We denote all these training examples as $\mathcal{D}_1$.
Similarly, for all the $t_0$ with $t_0\bmod 2=0$, their global minima are centering around $\theta_2^*=-\frac{1}{\delta}$. We denote all these training examples as $\mathcal{D}_2$.
In this toy dataset, the heterogeneity comes from the fact that $\mathcal{D}_1$ and $\mathcal{D}_2$ are not gathering around the same global minimia:  when $\delta$ increases, the distance of two minima $|\delta+1/\delta|$ centers will increase.
We now present the problem with uniform sampling in the following lemma.

\begin{lemma}\label{lemma heterogeneity noise}
Consider using uniform sampling on $\mathcal{D}$ to obtain a mini-batch $\xi$ with size of $M=2$,
    then when the training examples in $\xi$ are either both sampled from $\mathcal{D}_1$ or both from $\mathcal{D}_2$, with high probability,
\begin{align}\label{heterogeneity noise}
    \text{Var}\left[\nabla f_{\xi}(\*\theta^{(0)})\right]=\underbrace{O(\delta^2)}_{\text{heterogeneity noise}} + o(\delta).
\end{align}
On the otherhand, when the training examples in $\xi$ are sampled from both $\mathcal{D}_1$ and $\mathcal{D}_2$ using stratified sampling, with high probability,
\begin{align}
    \text{Var}\left[\nabla f_{\xi}(\*\theta^{(0)})\right]=\underbrace{o(\delta)}_{\text{model randomness}}.
\end{align}
\end{lemma}
%From this simple derivation we can see that in uniform sampling, when samples come from the same or close distributions, the batch fails to capture the diversity of the gradients, and thus will suffer from additional heterogeneity noise -- a result, as we mentioned before, comes from the non-identical distribution of this time series. 
%In other words, if we adopt a different sampler where $\mathcal{E}$ does not happen, then the gradient estimator will not contain the heterogeneity error with better estimation accuracy.
Lemma~\ref{lemma heterogeneity noise} shows with uniform sampling, the variance of the stochastic gradients is closely related to the samples: If the samples are somewhat similar, the variance on the gradient will suffer from an additional term -- to which we informally refer as heterogeneity noise.

\textbf{Proof to Lemma~\ref{lemma heterogeneity noise}.}
\begin{proof}
Note that for AR(1), the model only has a single parameter. In this proof, $\*\theta$ and $[\theta]$ are interchangeable.
The loss functions at different time can be expressed as:
\[f_{1,t}(\*\theta)=f_{1,t}([\theta])=\begin{cases}
    (\delta\theta+1+\epsilon_t)^2 & t\bmod 4=0\\
      (\theta-\delta-\epsilon_t)^2 & t\bmod 4=1 \\
      (\delta\theta+1-\epsilon_t)^2 & t\bmod 4=2 \\
      (\theta-\delta+\epsilon_t)^2 & t\bmod 4=3 
   \end{cases},
\]
Take gradient with respect to the model parameter $\theta$ and without the loss of generality, taking $\*\theta^{(0)}=[0]$, we obtain
\begin{align}
\nabla f_{1,t}([0])=\begin{cases}
    2\delta+2\delta\epsilon_t & t\bmod 4=0\\
      -2\delta-2\epsilon_t & t\bmod 4=1 \\
      2\delta-2\delta\epsilon_t & t\bmod 4=2 \\
      -2\delta+2\epsilon_t & t\bmod 4=3
   \end{cases}.
\end{align}
As defined by Equation~(\ref{equation_objective}), the gradient on the total loss can be expressed as (with out the loss of generality, we set $T=4\tilde{T}$ where $\tilde{T}$ is an integer.)
\begin{align}
    \nabla f(\*\theta^{(0)})=\frac{1}{T}\sum_{t=1}^{T}\nabla f_{1,t}(\*\theta^{(0)})=\frac{2}{T}\sum_{m=0}^{\tilde{T}-1}(-\epsilon_{m+1}-\delta\epsilon_{m+2}+\epsilon_{m+3}+\delta\epsilon_{m+4}).
\end{align}
Denote $\mathcal{E}$ as the event of "two training examples in the mini-batch are either both sampled from $\mathcal{D}_1$ or both sampled from $\mathcal{D}_2$". Depend on the event of $\mathcal{E}$, we first obtain when the event $\mathcal{E}$ happens, 
\begin{align}
    & \text{Var}\left[\nabla f_{\xi^{(0)}}(\*\theta^{(0)})\Big|\mathcal{E}\right]\\
    = & \frac{1}{T^2}\sum_{t,t'\bmod 4 = 0}\left|2\delta + \delta\epsilon_t + \delta\epsilon_{t'} - \frac{2}{T}\sum_{m=0}^{\tilde{T}-1}(-\epsilon_{m+1}-\delta\epsilon_{m+2}+\epsilon_{m+3}+\delta\epsilon_{m+4})\right|^2\\
        & + \frac{1}{T^2}\sum_{t,t'\bmod 4 = 1}\left|-2\delta - \epsilon_t - \epsilon_{t'} - \frac{2}{T}\sum_{m=0}^{\tilde{T}-1}(-\epsilon_{m+1}-\delta\epsilon_{m+2}+\epsilon_{m+3}+\delta\epsilon_{m+4})\right|^2\\
        & + \frac{1}{T^2}\sum_{t,t'\bmod 4 = 2}\left|2\delta - \delta\epsilon_t - \delta\epsilon_{t'} - \frac{2}{T}\sum_{m=0}^{\tilde{T}-1}(-\epsilon_{m+1}-\delta\epsilon_{m+2}+\epsilon_{m+3}+\delta\epsilon_{m+4})\right|^2\\
        & + \frac{1}{T^2}\sum_{t,t'\bmod 4 = 3}\left|-2\delta + \epsilon_t + \epsilon_{t'} - \frac{2}{T}\sum_{m=0}^{\tilde{T}-1}(-\epsilon_{m+1}-\delta\epsilon_{m+2}+\epsilon_{m+3}+\delta\epsilon_{m+4})\right|^2\\
    & + \frac{1}{T^2}\sum_{t\bmod 4 = 0,t'\bmod 4 = 2}\left|2\delta + \delta\epsilon_t - \delta\epsilon_{t'} - \frac{2}{T}\sum_{m=0}^{\tilde{T}-1}(-\epsilon_{m+1}-\delta\epsilon_{m+2}+\epsilon_{m+3}+\delta\epsilon_{m+4})\right|^2\\
        & + \frac{1}{T^2}\sum_{t\bmod 4 = 2,t'\bmod 4 = 0}\left|2\delta - \delta\epsilon_t + \delta\epsilon_{t'} - \frac{2}{T}\sum_{m=0}^{\tilde{T}-1}(-\epsilon_{m+1}-\delta\epsilon_{m+2}+\epsilon_{m+3}+\delta\epsilon_{m+4})\right|^2\\
        & + \frac{1}{T^2}\sum_{t\bmod 4 = 1,t'\bmod 4 = 3}\left|-2\delta - \epsilon_t + \epsilon_{t'} - \frac{2}{T}\sum_{m=0}^{\tilde{T}-1}(-\epsilon_{m+1}-\delta\epsilon_{m+2}+\epsilon_{m+3}+\delta\epsilon_{m+4})\right|^2\\
        & + \frac{1}{T^2}\sum_{t\bmod 4 = 3,t'\bmod 4 = 1}\left|-2\delta + \epsilon_t - \epsilon_{t'} - \frac{2}{T}\sum_{m=0}^{\tilde{T}-1}(-\epsilon_{m+1}-\delta\epsilon_{m+2}+\epsilon_{m+3}+\delta\epsilon_{m+4})\right|^2\\
    % triangle inequality
    \leq & \frac{4}{T^2}\sum_{t,t'\bmod 4 = 0}\delta^2 + \frac{1}{T^2}\sum_{t,t'\bmod 4 = 0}\left|\delta\epsilon_t + \delta\epsilon_{t'} - \frac{2}{T}\sum_{m=0}^{\tilde{T}-1}(-\epsilon_{m+1}-\delta\epsilon_{m+2}+\epsilon_{m+3}+\delta\epsilon_{m+4})\right|^2\\
        & \cdots \\
        & + \frac{4}{T^2}\sum_{t\bmod 4 = 3,t'\bmod 4 = 1}\delta^2 + \frac{1}{T^2}\sum_{t\bmod 4 = 3,t'\bmod 4 = 1}\left|\epsilon_t - \epsilon_{t'} - \frac{2}{T}\sum_{m=0}^{\tilde{T}-1}(-\epsilon_{m+1}-\delta\epsilon_{m+2}+\epsilon_{m+3}+\delta\epsilon_{m+4})\right|^2
\end{align}
where in the last step we apply $|a+b|^2\leq 2a^2+2b^2, \forall a, b\in\mathbb{R}$, and we break each term into two parts. The first part is independent of $\epsilon_t$ and is only related to $\delta^2$, and the second term is the average of some sequence of $\epsilon_t$. Then $\text{Var}\left[\nabla f_{\xi^{(0)}}(\*\theta^{(0)})\Big|\mathcal{E}\right]$ can be upper bounded by the following form
\begin{align}
    \text{Var}\left[\nabla f_{\xi^{(0)}}(\*\theta^{(0)})\Big|\mathcal{E}\right]\leq O(\delta^2) + \mathcal{T}_{\epsilon},
\end{align}
where $\mathcal{T}_{\epsilon}$ is the term containing all the $\epsilon_t$. Next we prove that $\mathcal{T}_{\epsilon}$ is a small quantity $o(\delta)$ with high probability.
Note that here all the $\epsilon_t$ are i.i.d. random variables, by applying the Hoeffding's inequality, which is for i.i.d random variables $Z_1, \cdots, Z_N$ following Gaussian white noise distribution,
\begin{align}
    \mathbb{P}\left(\left|\frac{1}{N}\sum_{m=1}^{N}Z_m\right|>t\right)\leq e^{-2Nt^2},
\end{align}
applying this to $\mathcal{T}_{\epsilon}$, we obtain (we show the derivation on the first term, the others are similar), with probability $1-e^{-2To(\delta)^2}$,
\begin{align}
    & \left|\delta\epsilon_t + \delta\epsilon_{t'} - \frac{2}{T}\sum_{m=0}^{\tilde{T}-1}(-\epsilon_{m+1}-\delta\epsilon_{m+2}+\epsilon_{m+3}+\delta\epsilon_{m+4})\right|^2\\
    \leq & \frac{1}{4}\left|\frac{1}{\tilde{T}}\sum_{m=0}^{\tilde{T}-1}(\epsilon_{m+1}-\epsilon_{m+3})\right|^2 + \frac{1}{4}\left|\frac{\delta}{\tilde{T}}\sum_{m=0}^{\tilde{T}-1}(\epsilon_{m+2}-\epsilon_{m+4} + \epsilon_t + \epsilon_{t'})\right|^2\\
    \leq & o(\delta).
\end{align}
Apply this to every term, we obtain with high probability,
\begin{align}
    \mathcal{T}_{\epsilon} \leq o(\delta).
\end{align}
We can do the similar analysis on $\text{Var}\left[\nabla f_{\xi^{(0)}}(\*\theta^{(0)})\Big|\neg\mathcal{E}\right]$, and obtain $\text{Var}\left[\nabla f_{\xi^{(0)}}(\*\theta^{(0)})\Big|\neg\mathcal{E}\right]\leq o(\delta)$. Here we omit this part for brevity.
\end{proof}

\subsection{Proof to Theorem~\ref{theorem_toyexample}}
\begin{proof}
Since this theorem states the existence of a dataset in order to show a lower bound,
the proof of is done by constructing such a dataset.
This implies
$\delta$ and $N$ can be chosen freely. Without the loss of generality, in the proof we treat the mini-batch $M=1$.

When $p=1$, we let $\mathcal{D}$ contains one single time series of length $p+1$ and $\delta>0$. It is straightforward to see the variance is zero (since the training here would be deterministic) and the theorem holds.

When $p\geq 2$,
the $\mathcal{D}$ we construct contains $N=2\left\lfloor p/2\right\rfloor$ different time series, and each time series is of length $p+1$. Note that here we have $N\geq 2$ because $p\geq 2$.
Since the model is predicting the same time here (it is predicting time $p+1$ given time $1$ to $p$), we let $c=\epsilon_p$ denote a fixed value\footnote{The $\epsilon_p$ can be obtained by generating the dataset using the same random seed as used in the model.}.
Without the loss of generality let $2c\leq\delta$ such that $\max_{i,t}|\*z_{i,t}|=\delta$.
Within the proof of this theorem, we let $\overline{p}=\left\lfloor p/2\right\rfloor$, $\mathcal{D}$ is constructed as the following:
\begin{align*}
    & \left[\frac{\delta}{2}, \underbrace{0, \cdots, 0}_{p-1}, \frac{\delta}{2}+c\right], i=1\\
    \text{Time Series i: } & \left[\underbrace{0, \cdots, 0}_{i-2}, \frac{\delta}{2}, -\frac{\delta}{2}, \underbrace{0, \cdots, 0}_{p-i}, c\right], 2\leq i\leq \overline{p}\\
    & \left[\underbrace{0, \cdots, 0}_{\overline{p}}, \underbrace{\frac{\delta}{2}\cdots, \frac{\delta}{2}}_{p-\overline{p}}, \frac{\delta}{2} + c\right], \overline{p}+1 \leq i\leq 2\overline{p}.
    %\text{Time Series $2\overline{p}+1$ to $2\overline{p}$: } & \left[\underbrace{0, \cdots, 0}_{\overline{p}}, \underbrace{\frac{\delta}{2}\cdots, \frac{\delta}{2}}_{p-\overline{p}}, c\right]
\end{align*}
Fit in the MSE loss we obtain:
\begin{align*}
    & f_1(\*\theta) = \left(\frac{\delta}{2} - \frac{\delta}{2}\*\theta_{1}\right)^2\\
    & f_i(\*\theta) = \left(\frac{\delta}{2}\*\theta_{i-1} - \frac{\delta}{2}\*\theta_{i}\right)^2, \forall 2\leq i\leq \overline{p}\\
    & f_i(\*\theta) = \left(\frac{\delta}{2} - \frac{\delta}{2}\*\theta_{\overline{p}+1} - \cdots - \frac{\delta}{2}\*\theta_{p}\right)^2, \forall \overline{p}+1 \leq i\leq 2\overline{p},
    %& f_i(\*\theta) = \left(\frac{\delta}{2}\*\theta_{\overline{p}+1} + \cdots + \frac{\delta}{2}\*\theta_{p}\right)^2, \forall 2\overline{p}+1 \leq i\leq 2\overline{p}
\end{align*}
where $f_i$ denotes the loss incurred on the $i$-th time series. The total loss function can then be expressed as
\begin{align}
    f(\*\theta) = & \frac{1}{2\overline{p}}\sum_{i=1}^{2\overline{p}}f_i(\*\theta) = \underbrace{\frac{1}{2\overline{p}}\sum_{i=1}^{\overline{p}}f_i(\*\theta)}_{g_1(\*\theta)} + \underbrace{\frac{1}{2\overline{p}}\sum_{i=\overline{p}+1}^{2\overline{p}}f_i(\*\theta)}_{g_2(\*\theta)}.
\end{align}
Note that when taking derivative of function $f$ with respect to the $\*\theta$, the first $\overline{p}$ coordinates will only be affected by $g_1(\*\theta)$, 
i.e.,
\begin{align}
    \frac{\partial f}{\partial\*\theta_i} = \frac{\partial g_1}{\partial\*\theta_i}, \forall i\leq \overline{p},
\end{align}
then we obtain
\begin{align}
    \|\nabla f(\*\theta)\|^2 = \sum_{j=1}^{p}\left|\frac{\partial f}{\partial\*\theta_i}\right|^2 \geq  \sum_{j=1}^{\overline{p}}\left|\frac{\partial f}{\partial\*\theta_i}\right|^2 =  \sum_{j=1}^{\overline{p}}\left|\frac{\partial g_1}{\partial\*\theta_i}\right|^2 = \|\nabla g_1(\*\theta)\|^2.
\end{align}
Lemma 1 in \cite{carmon2019lower} shows that for every $\*\theta$ with $\*\theta_{\overline{p}}=0$,
\begin{equation}
    \|\nabla g_1(\*\theta)\| \geq \frac{\delta^2}{4\overline{p}^{\frac{5}{2}}}.
\end{equation}
As a result, we obtain for every $\*\theta$ with $\*\theta_{\overline{p}}=0$,
\begin{equation}\label{equation_proof_toyexample_lbforf}
    \|\nabla f(\*\theta)\| \geq \frac{\delta^2}{4\overline{p}^{\frac{5}{2}}}.
\end{equation}
Without the loss of generality we set\footnote{If the initialization $\*v\neq\*0$, we just need to replace all the $\*\theta$ with $\*\theta-\*v$ in the original functions and the proof will be the same.} $\*\theta^{(0)}=\*0$. Now if we look at the expression of function $g_1$, if the model starts from $\*0$, it follows a "zero-respecting" property: $\*\theta_j$ will remain zero if fewer then $j$ number of gradients on $g_1$ is computed \cite{carmon2019lower}. 
Define a filtration $\mathcal{F}^{(t)}$ at iteration $t$ as the sigma field of all the previous events happened before iteration $t$.
Let $\tau_i$ denote the recent time where sample $i$ is sampled for computing the stochastic gradient. And let $N_t$ be a random variable, denoting the largest number where $\tau_1$ to $\tau_{N_t}$ is strictly increasing. Since each sample is uniformly sampled, we obtain
\begin{equation}
    \mathbb{P}[N^{(t+1)} - N^{(t)} = 1 | \mathcal{F}^{(t)}] \leq \frac{1}{2\overline{p}} \leq \frac{1}{\overline{p}}.
\end{equation}
Let $q^{(t)} = N^{(t+1)} - N^{(t)}$, with Chernoff bound, we obtain
\begin{equation}
    \mathbb{P}[N^{(t)} \geq {\overline{p}}] = \mathbb{P}[e^{\sum_{j=0}^{t-1}q^{(j)}} \geq e^{\overline{p}}] \leq e^{-{\overline{p}}}\mathbb{E}[e^{\sum_{j=0}^{t-1}q^{(j)}}].
\end{equation}
For the expectation term we know that
\begin{equation}
    \mathbb{E}[e^{\sum_{j=0}^{t-1}q^{(j)}}] = \mathbb{E}\left[\prod_{j=0}^{t-1}\mathbb{E}\left[e^{q^{(j)}}|\mathcal{F}^{(j)}\right]\right] \leq \left(1-\frac{1}{{\overline{p}}}+\frac{e}{{\overline{p}}}\right)^t \leq e^{t(e-1)/{\overline{p}}}.
\end{equation}
Thus we know
\begin{equation}
    \mathbb{P}[N^{(t)}\geq {\overline{p}}] \leq e^{\frac{(e-1)t}{{\overline{p}}}-{\overline{p}}} \leq \omega,
\end{equation}
for every $t\leq \frac{{\overline{p}}^2+{\overline{p}}\log(\omega)}{(e-1)}$.
Take $\omega=\frac{1}{2}$, for any $0<\epsilon<\frac{\delta^2}{8p^{\frac{5}{2}}}\leq\frac{\delta^2}{8\overline{p}^{\frac{5}{2}}}$, we obtain
\begin{align}
    \mathbb{E}\|\nabla f(\*\theta)\| = & \mathbb{P}(N^{(t)}\geq {\overline{p}})\left[\|\nabla f(\*\theta)\| \Big| N^{(t)}\geq {\overline{p}}\right] + \mathbb{P}(N^{(t)}< {\overline{p}})\left[\|\nabla f(\*\theta)\| \Big| N^{(t)}< {\overline{p}}\right]\\
    \geq & \frac{1}{2}\|\nabla f(\*\theta)\|\\
    > & \frac{1}{2}\cdot 2\epsilon\\
    = & \epsilon,
\end{align}
where we use Equation~(\ref{equation_proof_toyexample_lbforf}). 
The gradient is calculated as follows:
\begin{align*}
    \nabla f_1(\*0) = & \left[-\frac{\delta^2}{2}, \underbrace{0, \cdots, 0}_{p-1}\right]\\
    \nabla f_i(\*0) = & \left[\underbrace{0, \cdots, 0}_{p}\right], 2\leq i\leq \overline{p}\\
    \nabla f_i(\*0) = & \left[\underbrace{0, \cdots, 0}_{\overline{p}}, \underbrace{-\frac{\delta^2}{2}\cdots, -\frac{\delta^2}{2}}_{p-\overline{p}}\right], \overline{p}+1\leq i\leq 2\overline{p}\\
    \nabla f(\*0) = & \left[-\frac{\delta^2}{4\overline{p}}, \underbrace{0, \cdots, 0}_{\overline{p}-1}, \underbrace{-\frac{\delta^2}{4\overline{p}}\cdots, -\frac{\delta^2}{4\overline{p}}}_{p-\overline{p}}\right].
\end{align*}
The sampling variance in the first iteration: 
\begin{align}
    \text{Var}\left[\nabla f_{\xi^{(0)}}(\*\theta^{(0)})\right] = & \mathbb{E}_{i\sim[2\overline{p}]}\|\nabla f_i(\*0) - \nabla f(\*0)\|^2\\
    = & \frac{1}{2\overline{p}}\left[\left(\frac{2\overline{p}-1}{4\overline{p}}\right)^2\delta^4 + \frac{p-\overline{p}}{16\overline{p}^2}\delta^4\right] + \frac{\overline{p}-1}{2\overline{p}}\left[\frac{p-\overline{p}+1}{16\overline{p}^2}\delta^4\right] + \frac{\overline{p}}{2\overline{p}}\left[\frac{1}{16\overline{p}^2}\delta^4+(p-\overline{p})\left(\frac{2\overline{p}-1}{4\overline{p}}\right)^2\delta^4\right]\\
    \leq & \frac{3}{\overline{p}}\delta^4 + \overline{p}\delta^4\\
    \leq & \overline{p}^2,
\end{align}
where the last step holds when we let $\delta^4\leq \min\{\overline{p}^3/6, \overline{p}\}$.
Since for every $t\leq \frac{\overline{p}^2+\overline{p}\log(\omega)}{(e-1)}$, $\mathbb{E}\|\nabla f(\*\theta)\|\geq\epsilon$, the lower bound on the iterations (number of gradients to be computed) $T_b$ is
\begin{equation}
    T_b = \Omega\left(\overline{p}^2\right),
\end{equation}
which implies
\begin{equation}
    T_b = \Omega\left(\text{Var}\left[\nabla f_{\xi^{(0)}}(\*\theta^{(0)})\right]\right).
\end{equation}
Furthermore,
\begin{align}
    \text{Var}\left[\nabla f_{\xi^{(0)}}(\*\theta^{(0)})\right] = &  \frac{1}{2\overline{p}}\left[\left(\frac{2\overline{p}-1}{4\overline{p}}\right)^2\delta^4 + \frac{p-\overline{p}}{16\overline{p}^2}\delta^4\right] + \frac{\overline{p}-1}{2\overline{p}}\left[\frac{p-\overline{p}+1}{16\overline{p}^2}\delta^4\right] + \frac{\overline{p}}{2\overline{p}}\left[\frac{1}{16\overline{p}^2}\delta^4+(p-\overline{p})\left(\frac{2\overline{p}-1}{4\overline{p}}\right)^2\delta^4\right]\\
    \geq & \frac{p-\overline{p}}{2}\left(\frac{2\overline{p}-1}{4\overline{p}}\right)^2\delta^4\\
    = & \Omega(\delta^4p),
\end{align}
and thus we complete the proof.
\end{proof}

\subsection{Proof to Theorem~\ref{theorem scott}}
\subsubsection{Main Proof}
\begin{proof}
Take the expectation with respect to the sampling randomness in the inner loop for $\*v^{(t,k)}$, we obtain
\begin{align}
    \mathbb{E}_{\xi^{(t,k)}} \left[\*v^{(t,k)}\right] = \mathbb{E}_{\xi^{(t,k)}} \left[ \nabla f(\*\theta^{(t,k)};\xi^{(t,k)})-\nabla f(\*\theta^{(t,0)};\xi^{(t,k)})+\*g^{(t,0)} \right] = \nabla f(\*\theta^{(t,k)}) + \underbrace{\*g^{(t,0)} - \nabla f(\*\theta^{(t,0)})}_{=\*\zeta_t}
\end{align}
Due to $\*\zeta_t$, the main step for SCott ($\*v^{(t,k)}$) is a biased estimation of the true gradient $\nabla f\left(\*\theta^{(t,k)}\right)$. The challenge of the proof is to handle such biasedness.
The rest of our analysis largely follows the proof routine in SCSG-type methods \cite{babanezhad2015stopwasting,lei2017non,li2018simple}. 
We do not take credit for those analysis.

\citet{li2018simple} proposes a nice framework of analyzing stochastic control variate type algorithm, where in their framework, the control variate $\*g^{(t,0)}$ is computed via a randomly sampled mini-batch. This, as we discussed in the paper, can be seen as a special case of random grouping. The main difference of SCott is in bounding $\*\zeta_t$ since here the noise is not from the uniform sampling. 
For brevity, we summarize several lemmas from previous work. and focus on analyzing $\*\zeta_t$ in the main proof.

We summarize the main results in Lemma~\ref{proof scott lemma2}. We encourage readers to refer to \cite{li2018simple} for complete derivation for this part of results. 
From Lemma~\ref{proof scott lemma2} we obtain when $\alpha_tL = cB^{-\frac{2}{3}}$ (where $c$ is a numerical constant),
\begin{align}
    & \alpha_tB\left(2-\frac{2}{B}-2\alpha_tL-\frac{1}{1-\alpha_t^2L^2B-\alpha_t^3L^3B^2}\right)\mathbb{E}\|\nabla f(\*\theta^{(t,0)})\|^2\\
        \leq & 2\mathbb{E}(f(\*\theta^{(t-1,0)}) - f(\*\theta^{(t,0)})) + 2\alpha_tB\left(1+\alpha_tL+\frac{1}{B}\right)\mathbb{E}\|\*\zeta_t\|^2,
\end{align}
given
$\alpha_tL = cB^{-\frac{2}{3}}$,
we obtain
\begin{align}
    1-\alpha_t^2L^2B-\alpha_t^3L^3B^2\geq 1-B^{-\frac{1}{3}}c^2-c^3,
\end{align}
put it back together with $\alpha_tL = cB^{-\frac{2}{3}}$, we get,
\begin{align}
    & cB^{\frac{1}{3}}\left(2-\frac{2}{B}-2cB^{-\frac{2}{3}}-\frac{1}{1-B^{-\frac{1}{3}}c^2-c^3}\right)\mathbb{E}\|\nabla f(\*\theta^{(t,0)})\|^2\\
        \leq & 2L\mathbb{E}(f(\*\theta^{(t-1,0)}) - f(\*\theta^{(t,0)})) + 2cB^{\frac{1}{3}}\left(1+cB^{-\frac{2}{3}}+B^{-1}\right)\mathbb{E}\|\*\zeta_t\|^2,
\end{align}
select $c\leq \frac{1}{4}$ we can get,
\begin{align}
    & 2-\frac{2}{B}-2cB^{-\frac{2}{3}}-\frac{1}{1-B^{-\frac{1}{3}}c^2-c^3} \geq \frac{1}{4}\\
    & 1+cB^{-\frac{2}{3}}+B^{-1} \leq 1.35.
\end{align}
Fit in Lemma~\ref{proof scott lemma3}, we obtain
\begin{align}
    \mathbb{E}\|\nabla f(\*\theta^{(t,0)})\|^2 \leq \frac{8L\mathbb{E}(f(\*\theta^{(t-1,0)}) - f(\*\theta^{(t,0)}))}{cB^{\frac{1}{3}}} + 11\sum_{i=1}^{B}w_i^2\sigma_i^2.
\end{align}
Telescoping from $t=0$ to $T-1$, we obtain
\begin{align}
    \mathbb{E}\left\|\nabla f\left(\hat{\*\theta}^{(T)}\right)\right\|^2 \leq & \frac{8(f(\*0) - \inf_{\*\theta}f(\*\theta))L}{cB^{\frac{1}{3}}T} + 11\sum_{i=1}^{B}w_i^2\sigma_i^2\\
    = & O\left(\frac{(f(\*0) - \inf_{\*\theta}f(\*\theta))L}{B^{\frac{1}{3}}T} + \sum_{i=1}^{B}w_i^2\sigma_i^2\right),
\end{align}
given our selection on the value of $B$,
\begin{align}
\sum_{i=1}^{B}w_i^2\sigma_i^2 \leq O\left(\epsilon^2\right),
\end{align}
and then, with
\begin{align}
    T = O\left(\frac{(f(\*0) - \inf_{\*\theta}f(\*\theta))L}{B^{\frac{1}{3}}\epsilon^2}\right),
\end{align}
we obtain
\begin{align}
    \mathbb{E}\left\|\nabla f\left(\hat{\*\theta}^{(T)}\right)\right\| \leq & \sqrt{\mathbb{E}\left\|\nabla f\left(\hat{\*\theta}^{(T)}\right)\right\|^2}\\
    = & \sqrt{O\left(\frac{(f(\*0) - \inf_{\*\theta}f(\*\theta))L}{B^{\frac{1}{3}}T} + \sum_{i=1}^{B}w_i^2\sigma_i^2\right)}\\
    \leq & \epsilon.
\end{align}
Applying Lemma~\ref{proof scott lemma1}, the total number of stochastic gradient being computed can then be calculated as
\begin{align}
    \sum_{t=0}^{T-1}(B + \mathbb{E}[K_t]) = 2BT = O\left(\frac{\Delta LB^{\frac{2}{3}}}{\epsilon^2}\right),
\end{align}
when $B=|\mathcal{D}|$, then $\sigma_{|\mathcal{D}|}^2=0$, the total number of gradients to be computed is
\begin{align}
    O\left(\frac{\Delta L|\mathcal{D}|^{\frac{2}{3}}}{\epsilon^2}\right),
\end{align}
when $B\neq|\mathcal{D}|$, then put in
\begin{align}
    B = O\left(\frac{B\sum_{i=1}^{B}w_i^2\sigma_i^2}{\epsilon^2}\right),
\end{align}
the total number of gradients to be computed is
\begin{align}
    O\left(\frac{\Delta L\left(B\sum_{i=1}^{B}w_i^2\sigma_i^2\right)^{\frac{2}{3}}}{\epsilon^{\frac{10}{3}}}\right).
\end{align}
And thus we complete the proof.
\end{proof}

\subsubsection{Technical Lemma}

\begin{lemma}\label{proof scott lemma1} 
    (\cite{horvath2020adaptivity}, Lemma 1)
    Let $N\sim$ Geom$(\gamma)$, for $\gamma>0$. Then for any sequence $D_0, D_1, \cdots$ with $\mathbb{E}|D_N|\leq\infty$,
    \begin{align}
        \mathbb{E}(D_N-D_{N+1}) = \left( \frac{1}{\gamma} - 1\right)(D_0-\mathbb{E}D_N).
    \end{align}
\end{lemma}

\begin{lemma}\label{proof scott lemma2} 
(\cite{li2018simple}, Proof to Theorem 3.1)
when $\alpha_tL = cB^{-\frac{2}{3}}$,
\begin{align}
    & \alpha_tB\left(2-\frac{2}{B}-2\alpha_tL-\frac{1}{1-\alpha_t^2L^2B-\alpha_t^3L^3B^2}\right)\mathbb{E}\|\nabla f(\*\theta^{(t,0)})\|^2\\
        \leq & 2\mathbb{E}(f(\*\theta^{(t-1,0)}) - f(\*\theta^{(t,0)})) + 2\alpha_tB\left(1+\alpha_tL+\frac{1}{B}\right)\mathbb{E}\|\*\zeta_t\|^2.
\end{align}
\end{lemma}
\begin{proof}
    The proof can be established straightforwardly by considering the contant batch size case in \cite{li2018simple}.
\end{proof}

\begin{lemma}\label{proof scott lemma3} 
    \begin{align}
        \mathbb{E} \|\*\zeta_t\|^2 \leq \sum_{i=1}^{B}w_i^2\sigma_i^2.
    \end{align}
\end{lemma}
\begin{proof}

\begin{align}
    & \mathbb{E} \|\*\zeta_t\|^2\\
    = & \mathbb{E} \| \*g^{(t,0)} - \nabla f(\*\theta^{(t,0)})\|^2\\
    = & \mathbb{E} \left\| \sum_{i=1}^{B}\frac{|\mathcal{D}_i|\nabla f(\*\theta^{(t,0)};\xi_i^{(t)})}{|\mathcal{D}|} - \nabla f(\*\theta^{(t,0)}) \right\|^2\\
    = & \mathbb{E} \left\| \sum_{i=1}^{B}\frac{|\mathcal{D}_i|\nabla f(\*\theta^{(t,0)};\xi_i^{(t)})}{|\mathcal{D}|} - \sum_{i=1}^{B} \frac{|\mathcal{D}_i|}{|\mathcal{D}|} \nabla f(\*\theta^{(t,0)}) \right\|^2\\
    = & \sum_{i=1}^{B} w_i^2 \mathbb{E} \left\| \nabla f(\*\theta^{(t,0)};\xi_i^{(t)}) -  \nabla f(\*\theta^{(t,0)}) \right\|^2 + \sum_{i\neq i'} \mathbb{E} \langle \nabla f(\*\theta^{(t,0)};\xi_i^{(t)}) -  \nabla f(\*\theta^{(t,0)}), \nabla f(\*\theta^{(t,0)};\xi_{i'}^{(t)}) -  \nabla f(\*\theta^{(t,0)})\rangle \\
    \leq & \sum_{i=1}^{B}w_i^2\sigma_i^2
\end{align}
where in the final step, we use the property that $\xi_i^{(t)}$ is independent of $\xi_{i'}^{(t)}$.
\end{proof}

\end{document}